\documentclass[lettersize,journal]{IEEEtran}
\usepackage{amsmath,amsfonts}
\usepackage{algorithmic}
\usepackage{algorithm}
\usepackage{array}
\usepackage[caption=false,font=normalsize,labelfont=sf,textfont=sf]{subfig}
\usepackage{textcomp}
\usepackage{stfloats}
\usepackage{url}
\usepackage{verbatim}
\usepackage{graphicx}
\usepackage{cite}

\usepackage{makecell}
\usepackage{multirow}
\usepackage{arydshln}
\usepackage{enumitem}
\usepackage{dsfont}
\usepackage{cleveref}

\crefname{figure}{Fig.}{Figs.}
\crefname{table}{Table}{Tables}
\crefname{equation}{Eq.}{Eqs.}

\begin{document}

\title{Through the PRISM: Preference Representation in Intermediate States of Video Diffusion Models}

\author{Haoxuan~Wu, Lai-Man~Po,~Senior~Member,~IEEE, Mengyang~Liu, Kun~Li, Hongzheng~Yang, Wei~Liu,~Fellow,~IEEE % <-this % stops a space
% \thanks{Manuscript received April 24, 2026.}
\thanks{This work has been submitted to the IEEE for possible publication. Copyright may be transferred without notice, after which this version may no longer be accessible.}
\thanks{Haoxuan Wu, Lai-Man Po, and Kun Li are with the Department of Electrical Engineering, City University of Hong Kong, Hong Kong, China. (e-mail: haoxuanwu2-c@my.cityu.edu.hk; eelmpo@cityu.edu.hk; kunli25-c@my.cityu.edu.hk)}
\thanks{Mengyang Liu and Wei Liu are with Video Rebirth, Hong Kong, China. (e-mail: lmyleon2014@gmail.com; wl2223@columbia.edu)}
\thanks{Hongzheng Yang is with the Chinese University of Hong Kong, Hong Kong, China. (e-mail: hzyang22@cse.cuhk.edu.hk)}
}


% \markboth{IEEE Transactions on Circuits and Systems for Video Technology}%
% {Shell \MakeLowercase{\textit{Li et al.}}: Efficient Latent Super-Resolution for Video Generation}
\markboth{}{}


\maketitle

\begin{abstract}
Video reward models are central to aligning video diffusion models with human preferences, yet existing ones are external vision-language models that score clean, decoded pixels. Scoring an intermediate denoising state therefore requires either finishing the remaining denoising steps or decoding a noisy latent that the reward model cannot reliably interpret. We ask whether a video generator can itself act as the evaluator, discriminating human preferences directly from its noisy latents. We propose PRISM (Preference Representation in Intermediate States of Diffusion Models), a latent video reward model that keeps a pre-trained video diffusion transformer frozen, reads features from its early blocks, and maps them to a timestep-conditioned reward through a lightweight query-based aggregation head with only 21.8M trainable parameters. PRISM outperforms pixel-space reward models at high noise levels on two benchmarks while remaining competitive with 8B-parameter vision-language reward models on clean videos. This noise robustness enables early-stage Best-of-N sampling: selecting among candidates after only a few denoising steps reduces sampling cost by up to 10× and surpasses the strongest pixel-space baseline on each generator at 3–4× lower cost.
\end{abstract}


\section{Introduction}

\IEEEPARstart{T}{he} rapid evolution of Video Diffusion Transformers \cite{latte,cogvideox,wan} has transformed video generation, enabling the creation of high-fidelity, temporally coherent content. However, ensuring these models align with complex human preferences remains a challenge \cite{imagereward,hpsv2,videoscore}, and assessing the perceptual quality of AI-generated videos has become an active research topic in its own right \cite{lmvq,hveval_plus}. While Video Reward Models (VRMs) \cite{videoReward,unifiedreward,videoscore2} are widely used for guiding this alignment, current approaches face significant limitations when applied to advanced optimization paradigms such as Reinforcement Learning (RL) \cite{lift,videoReward} and Inference-Time Scaling \cite{vbs,vt1,vits1}.

Existing VRMs \cite{videoscore, videoReward, unifiedreward, videoscore2} operate in the pixel space, functioning as external evaluators built upon Vision-Language Models (VLMs). This conventional paradigm introduces a cascading series of bottlenecks rooted in a \textbf{Architectural Mismatch}. Because these VRMs are structurally distinct from the video diffusion backbones they evaluate, they are restricted to isolated, offline updates, sacrificing the joint scaling and self-evolution paradigm that has proven highly effective in LLMs \cite{ppo, instructgpt, llama2}. Furthermore, this architectural separation restricts evaluation exclusively to the clean pixel domain. Consequently, these models cannot interpret the intermediate, noisy latent states crucial for alignment strategies, such as step-level RL or early rejection in Best-of-$N$ sampling. Forcing evaluations into the pixel space by repeatedly decoding these noisy latents not only yields degraded visual signals that confuse the external VRMs, but also imposes a prohibitive computational burden \cite{vbs,vt1}.

\begin{figure}[tb]
  \centering
  \includegraphics[width=\columnwidth]{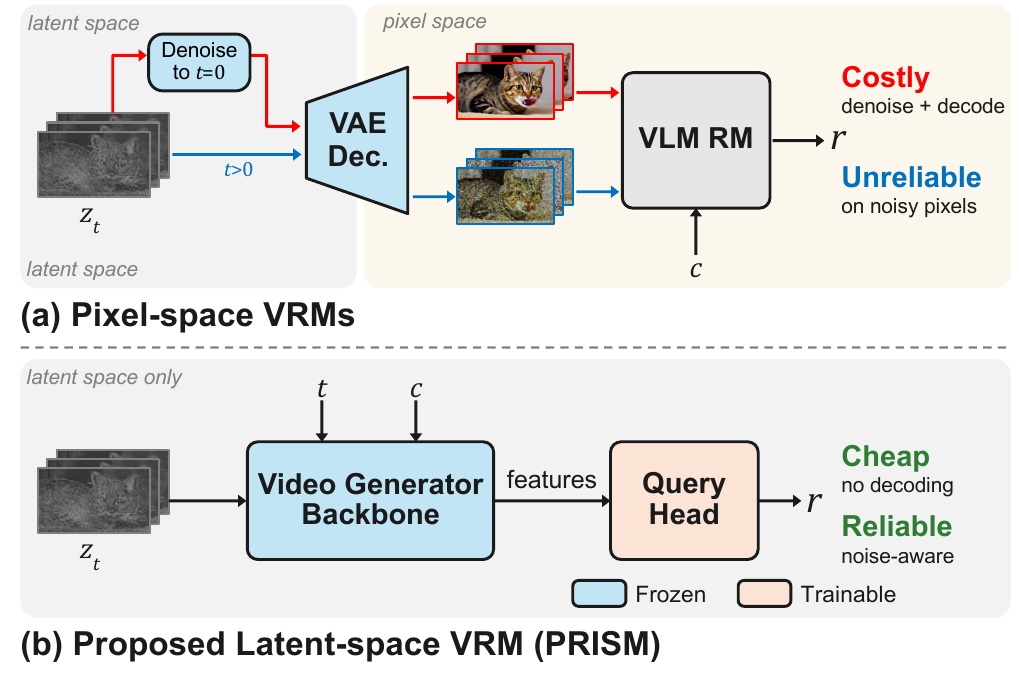}
    \caption{Comparison of video reward modeling paradigms. (a) Pixel-space VRMs score videos with an external VLM-based reward model, so a noisy latent $z_t$ must either be fully denoised and decoded (red), which is costly, or be decoded directly at $t>0$ (blue), yielding noisy pixels on which the reward is unreliable. (b) Our latent-space VRM, PRISM, reuses the frozen video generator backbone: its intermediate features, conditioned on timestep $t$ and prompt $c$, are aggregated by a lightweight query head into the reward $r$, requiring no VAE decoding and applying at any timestep.}
  \label{fig:introfig}
\end{figure}

These challenges motivate Latent Video Reward Modeling. We challenge the necessity of external evaluators by asking: \textit{Does a powerful video generator inherently possess the ability to discriminate human preferences, even when the visual content is severely obscured by noise?} Recent work (e.g., DDO \cite{ddo}) reveal that likelihood-based generative models secretly possess strong discriminative capabilities. Building on this, we posit that a pre-trained diffusion backbone is not merely a generator, but a rich storehouse of spatio-temporal priors. Its core training objective—reconstructing clean content from varying noise levels—equips it with an intrinsic \textit{blueprint} of the natural video manifold \cite{repa}. By repurposing the generator itself as a natively noise-aware evaluator operating within the latent space, we eliminate VAE decoding overheads, in line with recent evidence that operating directly in the compact latent space substantially reduces the cost of video synthesis \cite{lk_1}. This approach provides reliable guidance under heavy noise, and its accuracy improves with the backbone, opening a path toward generators that can evaluate and improve themselves.

\begin{figure*}[!t]
  \centering
  \includegraphics[width=\textwidth]{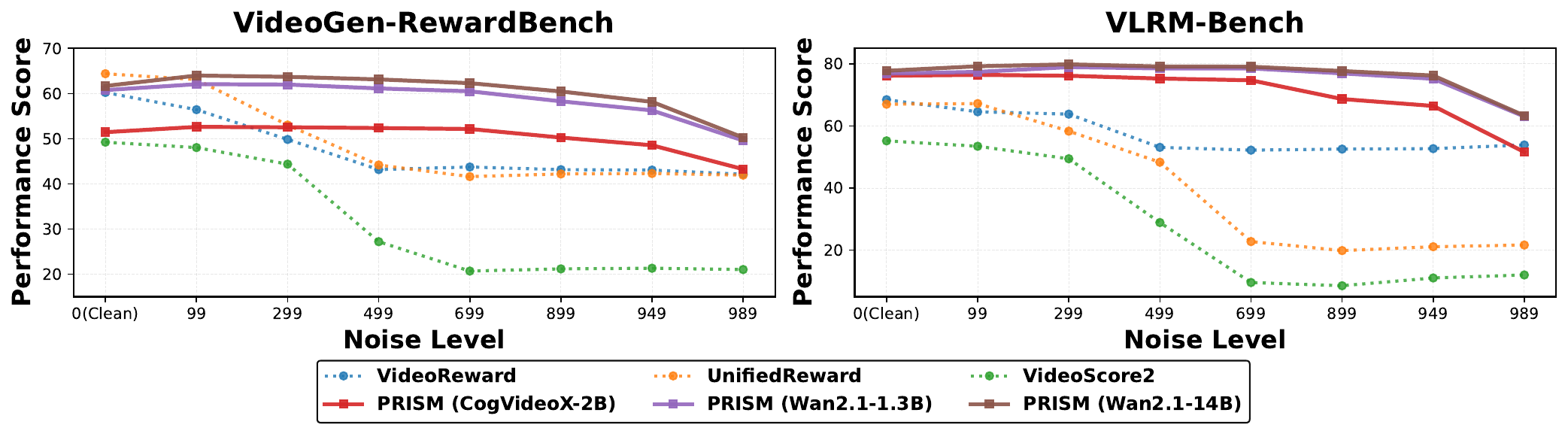}
  \caption{Preference accuracy (w/ Ties) across noise levels $t$ on (left) VideoGen-RewardBench and (right) VLRM-Bench. Pixel-level reward models (dotted lines) degrade sharply or even collapse as $t$ increases, whereas PRISM (solid lines) remains accurate throughout the denoising trajectory and generally outperforms the strongest pixel-level baselines, even with the 1.3B backbone.}
  \label{fig:bench_com}
\end{figure*}

Motivated by these observations, we introduce \textbf{PRISM} (\textbf{P}reference \textbf{R}epresentation in \textbf{I}ntermediate \textbf{S}tates of Diffusion \textbf{M}odels), as illustrated in \cref{fig:introfig}(b). Rather than resorting to expensive full-parameter fine-tuning \cite{lpo, prfl}, PRISM freezes the pre-trained video diffusion backbone. This design choice not only ensures training efficiency but preserves the backbone's intrinsic ability to interpret noisy video latents. Given that the frozen generator already captures video semantics, relying on a structurally redundant external VLM becomes unnecessary. The only remaining challenge is how to decode the implicit preference information from the backbone's high-dimensional, noise-corrupted intermediate features. To bridge this gap, we introduce a Query-based Aggregation head. Acting as a dedicated information extractor, it captures preference signals from the spatio-temporal features. By repurposing the generator's priors, this efficient architecture achieves the best alignment accuracy on noisy latents while exhibiting strong noise-robustness (\cref{fig:bench_com}).

Our main contributions are summarized as follows:
\begin{itemize}
    \item Decoding-free, Noise-aware Reward Framework. We introduce PRISM, a novel latent video reward model that completely freezes the generative backbone. By incorporating a Query-based Aggregation head, PRISM disentangles semantic preference signals from severe noise, avoiding the massive overhead of VAE decoding.
    
    \item Insights into Generative Priors and Evaluative Power. We provide the first systematic study demonstrating a strong positive correlation between a video diffusion backbone's generative capabilities and its inherent reward modeling potential. Our findings confirm that these generative priors remain informative under heavy noise.
    
    \item Noise-robust Preference Accuracy and Efficient Inference-time Scaling. PRISM achieves superior accuracy on noisy latents, outperforming pixel-space VRMs by large margins under heavy noise while staying competitive on clean videos. This enables early-stage Best-of-$N$ sampling that prunes inferior candidates within a few denoising steps, reducing sampling cost while generally improving generation quality.
\end{itemize}

\section{Related Work}

\subsection{Video Generation Models}
Text-to-video generation has rapidly evolved from U-Net-based diffusion architectures \cite{unet,xuetowards} toward scalable Diffusion Transformers (DiTs) \cite{DiT}. Recent large-scale models, including CogVideoX \cite{cogvideox}, HunyuanVideo \cite{hunyuanvideo}, SkyReels-V2 \cite{skyreelsv2}, and Wan \cite{wan}, increasingly perform generation in compact spatio-temporal latent spaces constructed by 3D causal VAEs, while employing Transformer backbones to model long-range spatial, temporal, and textual dependencies. CogVideoX adopts a 3D causal VAE together with an expert Transformer for text-video interaction, whereas HunyuanVideo and Wan further scale this latent video generation paradigm to large Transformer architectures. SkyReels-V2 additionally explores Diffusion Forcing to support long-duration video generation. Built on these latent DiT backbones, recent studies further explore customized subject-driven video generation \cite{tcsvt_2} and efficient high-resolution synthesis through lightweight latent upscaling \cite{lk_1}.

Despite their architectural similarities, these models do not adopt identical generative parameterizations. CogVideoX follows a diffusion formulation with velocity prediction, while recent systems such as HunyuanVideo and Wan employ Flow Matching or related flow-based formulations \cite{flow1}. Importantly, these models learn rich spatio-temporal representations throughout the denoising trajectory. PRISM leverages such internal representations to estimate human preferences directly from intermediate video latents.

\begin{figure*}[!t]
\centering
\includegraphics[width=\textwidth]{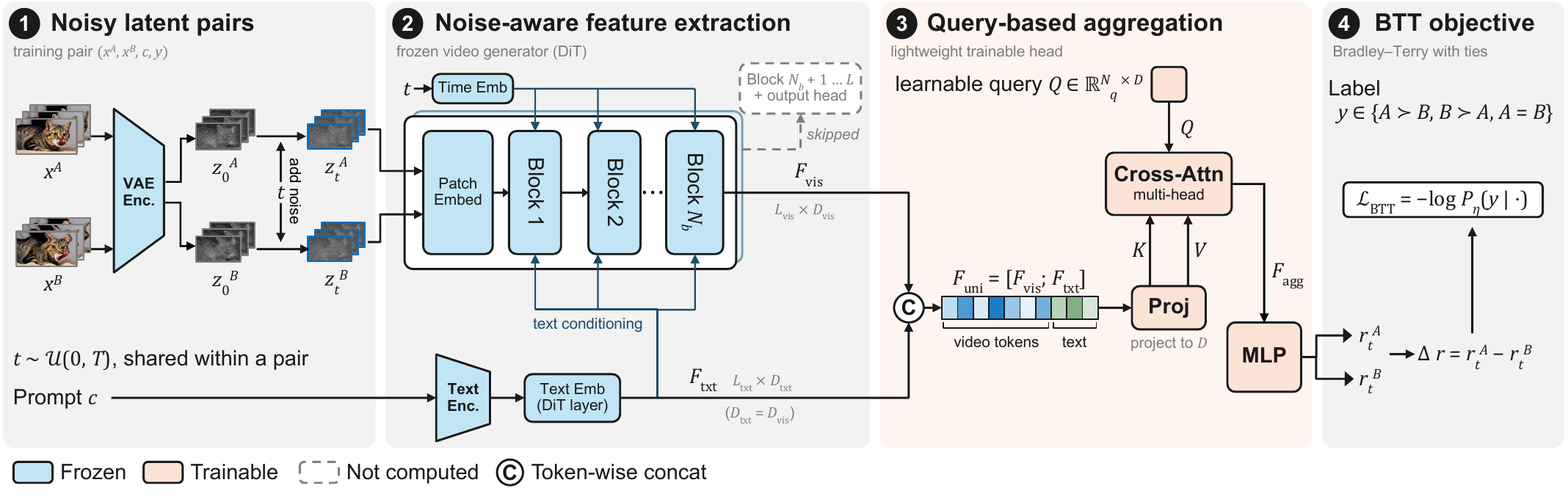}
    \caption{
    Overview of PRISM. (1) Both videos of an annotated pair $(x^A, x^B, c, y)$ are encoded by the frozen VAE encoder and noised to a timestep $t \sim \mathcal{U}(0,T)$ shared within the pair. (2) The frozen video generator (DiT), conditioned on $t$ and prompt $c$, executes only its first $N_b$ blocks on each noisy latent, yielding spatio-temporal features $F_{\mathrm{vis}}$; text features $F_{\mathrm{txt}}$ are taken from its own text-embedding layer. (3) The trainable query-based aggregation head concatenates them into $F_{\mathrm{uni}}$, over which a learnable query $Q$ cross-attends to obtain $F_{\mathrm{agg}}$, mapped by an MLP to a scalar reward. (4) The pair rewards $r_t^A, r_t^B$ are optimized with the Bradley–Terry-with-ties loss $\mathcal{L}_{\mathrm{BTT}}$ (tie threshold $\eta$).
    }
\label{fig:method}
\end{figure*}

\subsection{Video Reward Models}
Video Reward Models (VRMs) provide preference signals for evaluating and aligning video generation models. Recent approaches increasingly build upon large Vision-Language Models (VLMs) \cite{mantis,Qwen2-VL,Qwen2.5-VL,llavaov}, enabling evaluation beyond conventional perceptual metrics. VideoReward \cite{videoReward}, built upon Qwen2-VL, models human preferences in terms of visual quality, motion quality, and text--video alignment, and further applies the learned reward to video generation alignment. UnifiedReward \cite{unifiedreward} develops a unified reward framework across multimodal understanding and generation tasks, supporting both pairwise preference comparison and pointwise scoring. VideoScore2 \cite{videoscore2}, based on Qwen2.5-VL, further performs multi-dimensional video evaluation and learns interpretable scoring through supervised fine-tuning and reinforcement learning.

Despite their strong performance on completed videos, these methods operate in the pixel domain. Evaluating an intermediate diffusion state therefore requires VAE decoding before the video can be processed by the VLM-based reward model. Moreover, such models are primarily trained on clean videos rather than noisy intermediate states, resulting in degraded preference accuracy as the diffusion timestep increases, as shown in \cref{fig:bench_com}. In contrast, PRISM predicts preferences directly from noisy video latents using representations from a pre-trained diffusion backbone, avoiding repeated decoding and explicitly supporting evaluation across different noise levels.

\subsection{Inference-Time Scaling}
Inference-time scaling improves generation quality by allocating additional computation during sampling rather than retraining the generator \cite{vits1,diff_its}. Best-of-$N$ (BoN) sampling is a representative strategy that generates multiple candidates and selects the one receiving the highest reward. Reward-guided selection and refinement have also been adopted in related generation tasks: Korkmaz et al. \cite{korkmaz_vlmsr} employ VLMs to select the most trustworthy output among multiple stochastic diffusion super-resolution samples, and BridgeIV \cite{tcsvt_2} applies test-time reward optimization to refine fine-grained details in customized video generation. While simple and effective, BoN becomes particularly expensive for video diffusion models, since every candidate typically requires a complete denoising trajectory followed by VAE decoding before pixel-based VRMs such as VideoReward \cite{videoReward} and VideoScore2 \cite{videoscore2} can evaluate its quality. Consequently, both generation and evaluation costs grow approximately proportionally with the number of sampled candidates.

A more efficient alternative is to reject low-quality trajectories before completing the full denoising process. Conventional VRMs are unsuitable for this purpose because intermediate latents must first be decoded, while their prediction reliability also decreases at higher noise levels. PRISM instead evaluates noisy latents directly, enabling candidate trajectories to be ranked at an early denoising stage. Low-reward candidates can then be terminated while computation is concentrated on promising trajectories, reducing the number of full denoising and decoding operations required by video BoN.

\section{Method}

In this section, we introduce PRISM (\textbf{P}reference \textbf{R}epresentation in \textbf{I}ntermediate \textbf{S}tates of Diffusion \textbf{M}odels), a reward model specifically designed to capture human preferences throughout the entire diffusion denoising trajectory. An overview of the framework is illustrated in \cref{fig:method}.

\subsection{Preliminaries}

Recent video generative models \cite{opensora, seedance1, hunyuanvideo} typically operate in a compressed latent space to alleviate computational burdens. Given a video $x$, a pretrained encoder $\mathcal{E}$ maps it into a latent representation $z_0 = \mathcal{E}(x) \in \mathbb{R}^{\mathcal{F} \times H \times W \times C}$, where $\mathcal{F}, H, W$, and $C$ denote the number of frames, latent height, latent width, and channel dimension, respectively. The generative process defines a forward trajectory that progressively transforms $z_0$ into Gaussian noise. Following a unified formulation, a noisy latent $z_t$ at timestep $t \in [0, T]$ can be sampled directly as:
\begin{equation}
z_t = \alpha_t z_0 + \sigma_t \epsilon, \qquad \epsilon \sim \mathcal{N}(\mathbf{0}, \mathbf{I}),
\label{eq:noisy}
\end{equation}
where $\alpha_t$ and $\sigma_t$ are time-dependent coefficients defining the probability path. In this convention, $t=0$ corresponds to the clean latent (where $\alpha_0=1, \sigma_0=0$), while $t=T$ indicates the maximum noise level. For standard diffusion, $\alpha_t = \sqrt{\bar{\alpha}_t}$ and $\sigma_t = \sqrt{1 - \bar{\alpha}_t}$; for flow matching frameworks, $\alpha_t$ and $\sigma_t$ typically follow a linear interpolation (e.g., $\alpha_t = 1 - t/T$). Video Diffusion Transformers are trained to reverse this process by learning a network $\mu_\theta(z_t, c, t)$ that predicts the added noise or the velocity field, conditioned on the text prompt $c$ and timestep $t$.

Specifically, for each timestep $t \in \{0, \dots, T\}$ in the forward process, we take $(z_t, c, t)$ as input and learn a time-conditioned reward function $r(z_t, c, t) \in \mathbb{R}$ from the frozen backbone's intermediate spatio-temporal representations. The reward is computed per-timestep from $(z_t, c, t)$ alone, enabling preference evaluation at arbitrary noise levels without requiring the full denoising trajectory.

\begin{figure}[tb]
  \centering
  \includegraphics[width=\columnwidth]{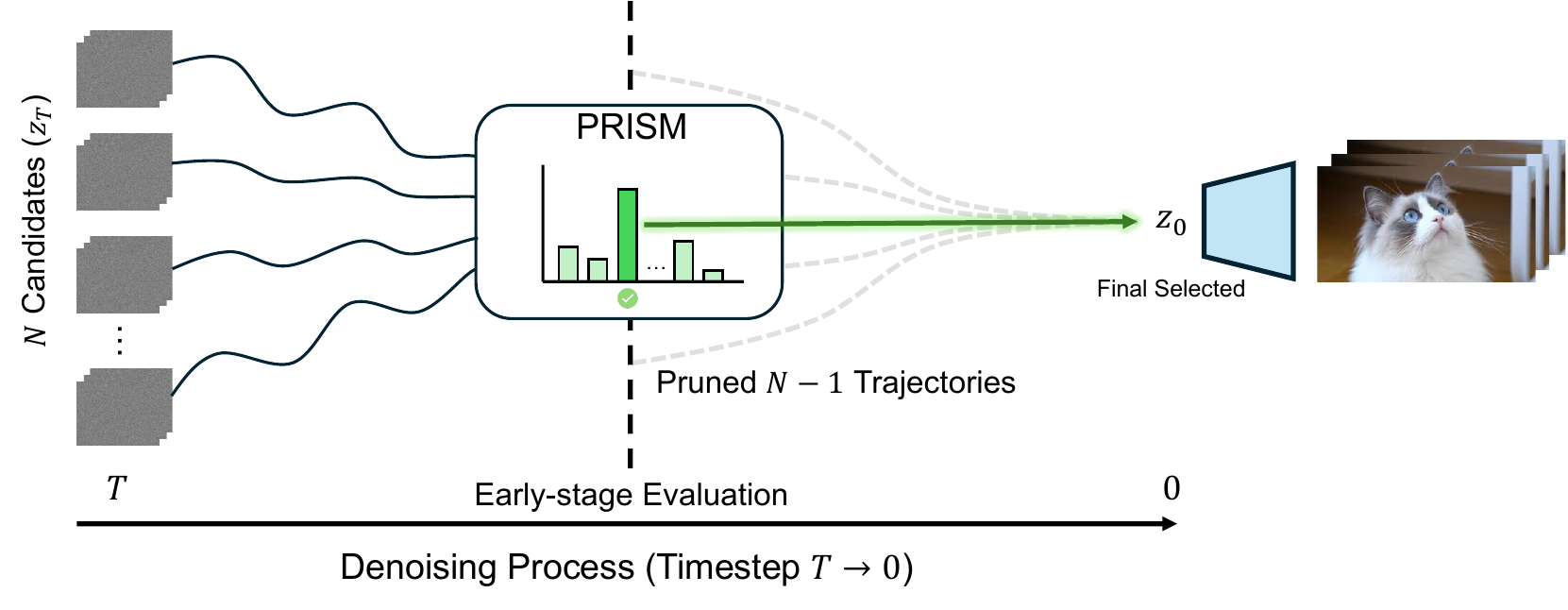}
        \caption{Best-of-$N$ (BoN) sampling pipeline empowered by PRISM. At an intermediate timestep, PRISM evaluates high-noise latents to identify the optimal candidate. The remaining $N-1$ suboptimal trajectories are immediately pruned, allowing only the single selected latent to complete the forward pass to the final pixel space.}
  \label{fig:BoN1}
\end{figure}

\subsection{Latent Video Reward Modeling} 

\subsubsection{Noise-aware Feature Extraction}

PRISM directly leverages the internal representations of a pre-trained Video Diffusion Transformer to construct a noise-aware evaluator. By repurposing the frozen generative backbone, we harness its inherent ability to capture complex spatio-temporal semantics and structural integrity across varying noise levels. 

Formally, given a noisy video latent $z_t$ and a text prompt $c$, we perform a single forward pass through the first $N_b$ blocks of the frozen diffusion backbone (stage 2 in \cref{fig:method}). This yields a set of intermediate spatio-temporal features $F_{\text{vis}} \in \mathbb{R}^{L_{\text{vis}} \times D_{\text{vis}}}$, computed as 

\begin{equation}
F_{\text{vis}} = \Phi_{\text{DiT}}(z_t, c, t).
\end{equation}

By extracting features at these intermediate layers rather than the final output layer, we capture low-level motion dynamics and high-level semantic alignment before they are entirely mapped to the denoising noise prediction. This strategy ensures a discriminative representation that maintains its robustness even at high noise levels ($t \to T$).

 % (i.e., $D_{txt} = D_{vis}$)
To maintain domain consistency, we employ the text encoder \cite{t5,umt5} in conjunction with the backbone's internal text embedding layer to extract textual features $F_{\text{txt}} \in \mathbb{R}^{L_{\text{txt}} \times D_{\text{txt}}}$. Since the text embedding is independent of the diffusion noise process, no enhancement is required. By deriving $F_{\text{txt}}$ from the backbone's embedding layer, it is aligned with $F_{\text{vis}}$ (where $D_{\text{txt}} = D_{\text{vis}}$ by construction), circumventing the need for additional projection layers at this stage.

\subsubsection{Feature Alignment and Aggregation}

While the frozen diffusion backbone provides robust, noise-resilient representations, the resulting spatio-temporal features $F_{\text{vis}}$ pose a challenge due to their length. Given the high resolution and temporal depth of video data, the sequence length $L_{\text{vis}}$ is often too large for direct processing. Without a proper bottleneck mechanism, this high-dimensional data leads to feature degradation, where preference signals (e.g., local motion artifacts or subtle distortions) are buried under redundant background tokens. A naive approach would be to employ global adaptive pooling \cite{lpo}. However, such reduction often exacerbates information loss, as it treats all tokens with equal importance, failing to capture fine-grained defects.

To mitigate this, we propose a Query-based Aggregation mechanism (stage 3 in \cref{fig:method}) designed to adaptively ``probe'' the feature sequence. We initialize a set of $N_q$ learnable queries $Q \in \mathbb{R}^{N_q \times D}$, which serve as information extractors to capture preference-relevant signals. Since the visual dimension $D_{\text{vis}}$ may vary across different backbones, we first concatenate $F_{\text{vis}}$ and $F_{\text{txt}}$, denoted as $F_{\text{uni}} = [F_{\text{vis}}; F_{\text{txt}}]$, and then apply a linear projection to map it into the unified dimension $D$. The queries then interact with the concatenated visual and textual features via a cross-attention mechanism \cite{transformer}:
\begin{equation}
F_{ \text{agg}} = \text{CrossAttn}(Q, K, V)
\end{equation}
where the keys $K \in \mathbb{R}^{(L_{\text{vis}} + L_{\text{txt}}) \times D}$ and values $V \in \mathbb{R}^{(L_{\text{vis}} + L_{\text{txt}}) \times D}$ are derived from the projected $F_{\text{uni}}$. This process allows the queries to dynamically attend to salient tokens across the entire video duration and spatial extent. In our implementation, we primarily set $N_q = 1$ to collapse the spatio-temporal tokens into a single concentrated global preference embedding $F_{\text{agg}}$, which is passed through an MLP to compute the scalar reward $r(z_t, c, t)$. Although average pooling is a standard baseline for feature aggregation, treating all positions in $F_{\text{vis}}$ and $F_{\text{txt}}$ equally yields sub-optimal performance. We provide a detailed discussion on this in the ablation section.

\subsection{Training Objectives}

PRISM is trained on a pairwise preference dataset $\mathcal{D}$. Each sample $(x^A, x^B, c, y) \in \mathcal{D}$ consists of a video pair $(x^A, x^B)$ generated from the same prompt $c$ and a ground-truth human preference label $y \in \{A \succ B,\ B \succ A,\ A = B\}$.

To ensure the model is noise-aware and capable of providing step-level guidance, we operate directly in the latent space. For each pair, we first encode the videos into clean latents $z_0^A = \mathcal{E}(x^A)$ and $z_0^B = \mathcal{E}(x^B)$ using the VAE of the diffusion backbone, and then perturb them into $z_t^A$ and $z_t^B$ at the same timestep $t$, shared within the pair, based on \cref{eq:noisy} (stage 1 in \cref{fig:method}). The reward model subsequently computes the scalar rewards $r_t^A = r(z_t^A, c, t)$ and $r_t^B = r(z_t^B, c, t)$ according to the architecture described in this section.

Given the inherent ambiguity in human perception, especially for videos of similar quality, we adopt the Bradley-Terry model with Ties (BTT) \cite{btt} (stage 4 in \cref{fig:method}) to formulate the preference probabilities. We introduce a tie-threshold parameter $\eta \ge 1$ to account for indifferent samples. For simplicity, let $E_A = \exp(r_t^A)$ and $E_B = \exp(r_t^B)$. The probabilities for each preference outcome are formulated as:

\begin{equation}
P_\eta(y \mid z_t^A, z_t^B, c, t) =
\begin{cases}
\frac{(\eta^2 - 1) E_A E_B}{(E_A + \eta E_B)(\eta E_A + E_B)}, & \text{if } A = B \\[1em]
\frac{E_A}{E_A + \eta E_B}, & \text{if } A \succ B \\[1em]
\frac{E_B}{\eta E_A + E_B}, & \text{if } B \succ A
\end{cases}
\end{equation}

The final training objective is to minimize the negative log-likelihood of the ground-truth preference labels across various noise levels $t$:
\begin{equation}
\mathcal{L}_{\text{BTT}} = - \mathbb{E}_{t \sim \mathcal{U}(0,T), (z^A, z^B, y, c) \in \mathcal{D}} \left[ \log P_\eta(y \mid z_t^A, z_t^B, c, t) \right]
\end{equation}
where the timestep $t$ is uniformly sampled from $\mathcal{U}(0, T)$. By optimizing this loss over the denoising trajectory, PRISM learns a robust and consistent preference metric. This noise-aware approach enables the model to bridge the gap between intermediate noisy latents and final clean outputs, providing reliable supervision for the alignment of video diffusion models.

\begin{table*}[!t]
\centering
\caption{Full quantitative results of preference prediction accuracy. We report performance across multiple benchmarks under various noise levels (timesteps $t$). Results are evaluated both with and without ties (``w/ Ties'' and ``w/o Ties''). For each evaluation setting, the best results are bolded, and the second-best results are underlined.}
\label{main_table_1}
   \begin{tabular}{ l c c c c c c c c }
      \hline
      \multirow{2}*{Model} & \multicolumn{8}{c}{Timestep ($t$)} \\
      ~ & 989 & 949 & 899 & 699 & 499 & 299 & 99 & 0(Clean) \\
      \hline
      \multicolumn{9}{l}{\textbf{VideoGen-RewardBench}} \\
      \multicolumn{9}{l}{w/ Ties} \\
      VideoReward & 42.13 & 43.05 & 43.14 & 43.73 & 43.16 & 49.83 & 56.43 & 60.23 \\
      UnifiedReward & 41.93 & 42.30 & 42.19 & 41.60 & 44.17 & 53.02 & \underline{63.10} & \textbf{64.39} \\
      VideoScore2 & 21.02 & 21.31 & 21.17 & 20.68 & 27.19 & 44.36 & 48.03 & 49.23 \\
      \hdashline
      PRISM (CogVideoX-2B) & 43.25 & 48.54 & 50.25 & 52.15 & 52.36 & 52.53 & 52.61 & 51.44 \\
      PRISM (Wan2.1-1.3B) & \underline{49.60} & \underline{56.26} & \underline{58.28} & \underline{60.50} & \underline{61.13} & \underline{61.99} & 62.07 & 60.76 \\
      PRISM (Wan2.1-14B) & \textbf{50.25} & \textbf{58.16} & \textbf{60.46} & \textbf{62.30} & \textbf{63.13} & \textbf{63.70} & \textbf{63.98} & \underline{61.68} \\
      \hline
      
      \multicolumn{9}{l}{w/o Ties} \\
      VideoReward & 50.64 & 51.74 & 51.84 & 52.55 & 51.87 & 59.88 & 67.81 & 72.38 \\
      UnifiedReward & 49.86 & 50.30 & 50.19 & 49.37 & 52.99 & 63.72 & \underline{75.83} & \textbf{77.38} \\
      VideoScore2 & 17.20 & 17.82 & 17.87 & 16.86 & 26.49 & 52.89 & 56.76 & 58.27 \\
      \hdashline
      PRISM (CogVideoX-2B) & 51.50 & 58.24 & 60.31 & 62.64 & 62.89 & 63.08 & 63.19 & 61.80 \\
      PRISM (Wan2.1-1.3B) & \underline{59.56} & \underline{67.56} & \underline{70.01} & \underline{72.68} & \underline{73.43} & \underline{74.42} & 74.51 & 72.93 \\
      PRISM (Wan2.1-14B) & \textbf{60.36} & \textbf{69.87} & \textbf{72.63} & \textbf{74.79} & \textbf{75.78} & \textbf{76.44} & \textbf{76.81} & \underline{74.12} \\
      \hline
      \hline
      \multicolumn{9}{l}{\textbf{VLRM-Bench}} \\
      \multicolumn{9}{l}{w/ Ties} \\
      VideoReward & 53.89 & 52.71 & 52.56 & 52.22 & 53.12 & 63.82 & 64.58 & 68.47 \\
      UnifiedReward & 21.66 & 21.11 & 19.86 & 22.77 & 48.33 & 58.33 & 67.22 & 67.01 \\
      VideoScore2 & 12.01 & 11.04 & 8.54 & 9.58 & 28.89 & 49.44 & 53.47 & 55.21 \\
      \hdashline
      PRISM (CogVideoX-2B) & 51.60 & 66.46 & 68.68 & 74.72 & 75.28 & 76.18 & 76.46 & 76.18 \\
      PRISM (Wan2.1-1.3B) & \underline{63.06} & \underline{75.21} & \underline{76.94} & \underline{78.54} & \underline{78.54} & \underline{78.96} & \underline{77.43} & \underline{76.88} \\
      PRISM (Wan2.1-14B) & \textbf{63.33} & \textbf{76.25} & \textbf{77.71} & \textbf{79.10} & \textbf{79.17} & \textbf{79.86} & \textbf{79.24} & \textbf{77.78} \\
      \hline
      \multicolumn{9}{l}{w/o Ties} \\
      VideoReward & 54.53 & 53.34 & 53.26 & 52.91 & 53.83 & 64.58 & 65.35 & 69.29 \\
      UnifiedReward & 21.43 & 20.81 & 19.53 & 22.55 & 48.91 & 59.03 & 68.02 & 67.81 \\
      VideoScore2 & 18.55 & 20.03 & 19.82 & 16.51 & 33.38 & 51.23 & 56.43 & 56.78 \\
      \hdashline
      PRISM (CogVideoX-2B) & 52.21 & 67.25 & 69.50 & 75.61 & 76.18 & 77.09 & 77.37 & 77.09 \\
      PRISM (Wan2.1-1.3B) & \underline{63.81} & \underline{76.11} & \underline{77.86} & \underline{79.48} & \underline{79.48} & \underline{79.90} & \underline{78.36} & \underline{77.86} \\
      PRISM (Wan2.1-14B) & \textbf{64.09} & \textbf{77.16} & \textbf{78.64} & \textbf{80.04} & \textbf{80.11} & \textbf{80.82} & \textbf{80.25} & \textbf{78.78} \\
      \hline
   \end{tabular}
\end{table*}

\section{Experiments}

\subsection{Implementation Details}

In our experiments, we utilize pre-trained text-to-video models as our default diffusion backbones, specifically CogVideoX-2B \cite{cogvideox}, Wan2.1-1.3B \cite{wan}, and Wan2.1-14B \cite{wan}. For each diffusion backbone, we extract features from the first 12 blocks. To ensure a fair comparison across different backbone architectures, we keep the aggregation head at a comparable scale for all backbones. The concatenated features are projected into the query dimension $D$ of the aggregation head, with $D=1536$ for both Wan2.1 backbones and $D=1920$, the native hidden size, for CogVideoX-2B. The aggregation employs a single learnable query ($N_q=1$), and the reward head consists of a 5-layer MLP. During training, the diffusion backbone remains frozen, and we only optimize the projection and aggregation modules. The BTT loss threshold $\eta$ is empirically set to 5.0. We employ the AdamW optimizer \cite{adamw} with learning rate of 1e-4.

Beyond the above training and evaluation settings for PRISM, we conduct a separate set of inference-time scaling experiments to evaluate its effectiveness when integrated with different video generation backbones. Specifically, we employ CogVideoX-2B and Wan2.1-1.3B as the generation models to further examine the architectural generalizability of our approach. For the Best-of-$N$ (BoN) sampling experiments, we set the candidate pool size to $N=5$. The evaluation is driven by the prompt suite provided by VBench \cite{vbench}, which spans multiple standard dimensions of video generation quality. To ensure alignment with official model capabilities and to avoid confounding variables introduced by ad-hoc prompt engineering, we utilize the enhanced prompt versions released by the VBench team. Furthermore, all inference hyperparameters strictly adhere to the official recommendations provided by the respective model developers and VBench evaluation guidelines. Specifically, for Wan2.1-1.3B, we generate videos at a spatial resolution of $832 \times 480$ with 81 frames at 16 FPS. For CogVideoX-2B, the outputs are configured to a resolution of $720 \times 480$, comprising 49 frames at 8 FPS.

\subsection{Experimental Settings}

\noindent \textbf{Training Dataset Construction and Annotation.}
We construct a large-scale pairwise preference dataset using videos generated by a diverse set of video generation models, including CogVideoX \cite{cogvideox}, OpenSora \cite{opensora}, HunyuanVideo \cite{hunyuanvideo}, the Wan2.1/2.2 series \cite{wan}, and SkyReels-V2 \cite{skyreelsv2}. To provide consistent textual conditions across different generators, we adopt the publicly released prompts from VBench \cite{vbench}. Each model is evaluated using its officially recommended inference configuration, including resolution, frame rate, number of frames, sampling steps, classifier-free guidance scale \cite{cfg}, and timestep shifting \cite{sd3}, where applicable. We do not introduce model-specific prompt engineering or additional enhancement techniques, thereby minimizing confounding factors and ensuring that the collected videos reflect the intrinsic generation capabilities of different models.

Each generated video pair is independently evaluated by three professional annotators along three complementary dimensions: \textit{Visual Quality}, \textit{Text Alignment}, and \textit{Motion Quality}. To improve annotation consistency, we establish detailed guidelines specifying the definition and evaluation criteria for each dimension, complemented by representative examples of common generation artifacts and pairwise preference cases. Before the main annotation stage, we conduct a pilot study on 1,000 video pairs to calibrate the annotators and refine the guidelines. During formal annotation, pre-collected ``golden pairs'' with expected outcomes are periodically inserted to monitor annotation quality, while regular calibration meetings are conducted to resolve ambiguous cases and maintain consistent evaluation standards.

To obtain reliable overall preference labels, we aggregate the dimension-level annotations using a conservative filtering strategy rather than simple majority voting. Specifically, a pair is retained only when one video strictly wins or ties with the other across all three dimensions: after removing dimension-level ``Tie'' judgments, all remaining preferences must consistently favor the same video. Pairs with conflicting preferences across dimensions are regarded as ambiguous and discarded, while pairs rated as ties on all dimensions are preserved as overall ties. This strategy filters out comparisons involving substantial quality trade-offs and retains high-confidence preference pairs. The resulting dataset contains 26,391 training pairs and 1,440 testing pairs. We enforce strict prompt-level disjointness between the training and test splits, and curate the latter as our standalone evaluation benchmark, termed \textbf{VLRM-Bench}. Compared with VideoGen-RewardBench, VLRM-Bench is built from more recent generators with fewer severe failures, so the two videos in a pair often differ only in subtle visual, motion, or alignment details.

\noindent \textbf{Evaluation Protocol and Metrics.}
We benchmark PRISM against representative video reward models, including VideoReward \cite{videoReward}, UnifiedReward \cite{unifiedreward}, and VideoScore2 \cite{videoscore2}. For fair comparison, all baselines are evaluated using their official checkpoints and recommended configurations. We first assess pairwise preference prediction accuracy on VideoGen-RewardBench \cite{videoReward}, reporting both ``w/ Ties'' and ``w/o Ties'' results following the established evaluation protocol. In addition, we evaluate on our curated VLRM-Bench, which contains preference pairs generated by recent video generation models with human-annotated labels, to examine generalization to unseen prompts. To further study robustness across the diffusion trajectory, we evaluate reward prediction at discrete noise levels rather than randomly sampling a timestep for each instance, allowing performance degradation with increasing noise to be analyzed in a controlled manner.

We further evaluate the practical utility of PRISM through inference-time scaling experiments based on Best-of-$N$ (BoN) sampling. Candidate videos are generated from VBench prompts \cite{vbench} with $N=5$. Conventional VLM-based reward models can only select candidates after full denoising and VAE decoding, whereas PRISM can score intermediate noisy latents at different denoising steps. This enables us to systematically evaluate the trade-off between early reward estimation, computational efficiency, and final generation quality. We report performance across the VBench Quality, Semantic, and Total scores and conduct experiments on both CogVideoX-2B and Wan2.1-1.3B to examine generalization across generation backbones. All generation settings follow the official model configurations and VBench evaluation guidelines.

\subsection{Experimental Results}

\subsubsection{Preference Prediction Accuracy}

We present the quantitative comparison of preference prediction in \cref{main_table_1}. To ensure a fair evaluation across the diffusion trajectory, pixel-level baselines are provided with videos reconstructed from noisy latents $z_t$ via the backbone's VAE decoder. This protocol reflects how pixel-space reward models are applied when only an intermediate latent is available. Estimating the clean latent with an additional forward pass of the generator before decoding would further increase their evaluation cost.

A key observation is the performance collapse of pixel-level models in high-noise scenarios. Specifically, VideoScore2 exhibits a \textit{tie-collapse} phenomenon: because it relies on an absolute scoring phase (mapping individual video to a discrete quality range), it tends to perceive all noisy inputs as \textit{complete failures} and assigns them the lowest possible score. This results in nearly all pairs being predicted as ``equal,'' leading to a catastrophic drop in accuracy at higher timesteps. Although UnifiedReward shows competitive results on low-noise samples in VideoGen-RewardBench, its performance drops as $t$ increases. In contrast, PRISM achieves the best accuracy on both benchmarks at all noisy timesteps ($t>0$) and remains competitive on clean videos. It preserves high accuracy even in high-noise cases where other methods fail, demonstrating the superior robustness of our noise-aware latent-level design. VLRM-Bench is more challenging for pixel-space reward models: on clean videos without ties, all three baselines score lower than on VideoGen-RewardBench, consistent with its higher overall video quality and smaller within-pair differences. PRISM outperforms all baselines at every timestep on VLRM-Bench, including clean videos.

Furthermore, \cref{main_table_1} compares PRISM variants using different diffusion backbones: CogVideoX-2B, Wan2.1-1.3B, and Wan2.1-14B. Two observations emerge from this comparison. \textbf{Intrinsic Quality vs. Parameter Scale.} Wan2.1-1.3B outperforms the CogVideoX-2B variant on V-bench, even though CogVideoX-2B has a larger parameter count and higher feature dimensionality. This suggests that the intrinsic representational capability of the backbone—likely stemming from its architecture or pre-training—is a more vital factor for reward modeling than raw model scale. \textbf{Scaling Dividends.} Within the same model family, scaling provides clear benefits. The Wan2.1-14B version consistently surpasses the 1.3B version, leveraging its larger hidden capacity and richer feature space for more precise preference distillation.

\begin{table*}[!t]
\centering
\caption{Full quantitative results for Best-of-$N$ (BoN) sampling. The ``Settings'' column specifies the denoising step at which PRISM performs selection. $\Delta$ denotes the performance gain. For each model, the best and second-best results are highlighted.}
\label{main_table_2}
   \begin{tabular}{ c c | c c c c }
      \hline
      \multirow{2}*{\textbf{RM}} & \multirow{2}*{\textbf{Settings}} & \multicolumn{4}{c}{\textbf{VBench}} \\
      ~ & ~ & Quality & Semantic & Total & $\mathrm{\Delta}$ \\
      \hline
      \multicolumn{6}{c}{\textbf{CogVideoX-2B}} \\
       - & - & 81.0631 & 77.0937 & 80.2693 & - \\
      \cline{1-6}
      Random & - & 81.3808 & 77.2381 & 80.5522 &  +0.2829 \\
      VideoReward & - & 81.6803 & 78.7097 & \underline{81.0862} &  \underline{+0.8169} \\
      UnifiedReward & - & 81.6947 & 77.4733 & 80.8504 &  +0.5811 \\
      VideoScore2 & - & 81.2159 & 78.1815 & 80.6090 &  +0.3397 \\
      \cline{1-6}
      \multirow{5}*{\makecell{PRISM\\(CogVideoX-2B)}} & 1 & 81.3019 & 77.4076 & 80.5230 &  +0.2537 \\
      ~ & 5 & 81.6351 & 77.7337 & 80.8549 &  +0.5856 \\
      ~ & 10 & 82.0087 & 77.7561 & \textbf{81.1582} &  \textbf{+0.8889} \\
      ~ & 25 & 81.7414 & 77.5840 & 80.9099 &  +0.6406 \\
      ~ & 50 & 81.5411 & 77.7721 & 80.7873 &  +0.5180 \\
      \hline
      \hline

      \multicolumn{6}{c}{\textbf{Wan2.1-1.3B}} \\
      
      - & - & 85.2300 & 75.6500 & 83.3100 & - \\
      \cline{1-6}
      Random & - & 85.5736 & 76.0586 & 83.6706 &  +0.3606 \\
      VideoReward & - & 85.3138 & 76.9041 & 83.6318 &  +0.3218 \\
      UnifiedReward & - & 85.2754 & 76.3370 & 83.4878 &  +0.1778 \\
      VideoScore2 & - & 85.9198 & 75.7450 & 83.8849 &  +0.5749 \\
      \cline{1-6}
      \multirow{5}*{\makecell{PRISM\\(Wan2.1-1.3B)}} & 1 & 85.6257 & 76.7701 & 83.8546 &  +0.5446 \\
      ~ & 5 & 85.9211 & 76.5801 & 84.0529 &  +0.7429 \\
      ~ & 10 & 85.8620 & 76.0182 & 83.8932 &  +0.5832 \\
      ~ & 25 & 86.0783 & 76.1780 & 84.0983 &  +0.7883 \\
      ~ & 50 & 86.0589 & 76.5513 & 84.1574 &  +0.8474 \\
      \cline{1-6}
      \multirow{5}*{\makecell{PRISM\\(Wan2.1-14B)}} & 1 & 85.9822 & 76.4926 & 84.0843 &  +0.7743 \\
      ~ & 5 & 86.1617 & 76.1093 & 84.1512 &  +0.8412 \\
      ~ & 10 & 86.0889 & 75.9489 & 84.0609 &  +0.7509 \\
      ~ & 25 & 86.3384 & 76.6792 & \textbf{84.4065} &  \textbf{+1.0965} \\
      ~ & 50 & 86.0786 & 76.9515 & \underline{84.2532} &  \underline{+0.9432} \\
      \hline
   \end{tabular}
\end{table*}

\subsubsection{Inference-time Scaling (BoN)}
We present the quantitative results of the inference-time scaling experiments in \cref{main_table_2}. To ensure latent-space consistency, each generator is paired with PRISM models whose backbones share its VAE: PRISM (CogVideoX-2B) for CogVideoX-2B, and both PRISM (Wan2.1-1.3B) and PRISM (Wan2.1-14B) for Wan2.1-1.3B, since the two Wan2.1 models share the same VAE.

As shown in \cref{main_table_2}, PRISM achieves the highest VBench Total on both generators, compared with the strongest pixel-space baselines. While VLM-based reward models are competitive on CogVideoX-2B, their gains shrink on the more recent Wan2.1-1.3B, where VideoReward and UnifiedReward even fall below the Random baseline. As a reference, we include a Random baseline that uniformly selects one of the N candidates. Since its expected score equals that of the base model, its gain of about 0.3 points reflects the sampling variance of VBench at N = 5, and differences of this magnitude should be interpreted with caution. For an intuitive overview of how our efficient selection mechanism operates, we visualize the complete PRISM sampling pipeline in \cref{fig:BoN1}.

\begin{figure*}[!t]
  \centering
  \includegraphics[width=\textwidth]{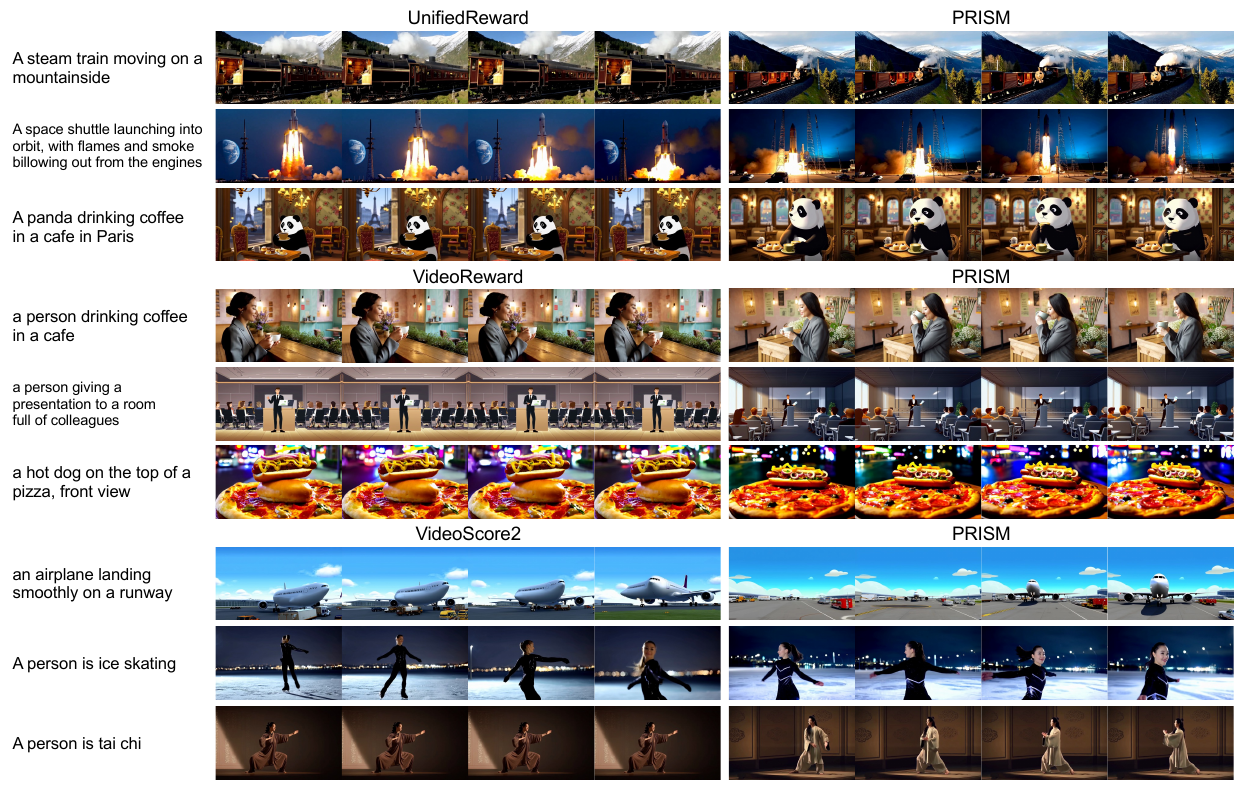}
\caption{Qualitative comparison of BoN results. Under identical prompts, PRISM identifies samples with semantic fidelity and physical consistency compared to pixel-based baselines (UnifiedReward, VideoReward, and VideoScore2). PRISM excels in capturing precise subject composition and articulated motion, which are often compromised in baseline-guided selections.}
  \label{fig:BoN_comp}
\end{figure*}

Beyond quantitative gains, the visual comparisons in \cref{fig:BoN_comp} further highlight PRISM's discriminative power. While baselines often suffer from subject counting artifacts, physically implausible motion, or a tendency to select near-static videos with minimal dynamic content (e.g., the fourth and ninth rows in \cref{fig:BoN_comp}), PRISM selects samples that adhere to semantic and physical constraints. Notably, PRISM is sensitive to fine-grained dynamics, such as the articulated hand movements of the panda (third row in \cref{fig:BoN_comp}), which are frequently overlooked by pixel-based evaluators.

\subsection{Efficiency Analysis}

\subsubsection{Parameter Efficiency}
Beyond inference-time acceleration, PRISM maintains a lightweight reward modeling architecture. As detailed in \cref{tab:param_scale}, existing pixel-based VRMs such as UnifiedReward and VideoScore2 rely on large VLM backbones with approximately 8B parameters, introducing substantial computational overhead. In contrast, PRISM directly leverages the pre-trained latent representations of the generative DiT backbone. By extracting features only from early DiT blocks while keeping the backbone completely frozen, PRISM reduces the operational parameter scale to as low as 604.76M. Moreover, the proposed query-based aggregation and reward head introduce only approximately 21.8M trainable parameters across all variants, accounting for less than 4\% of the smallest backbone configuration. This lightweight design substantially reduces the optimization overhead required for reward modeling and facilitates deployment with limited computational resources.

\subsubsection{Inference-Time Efficiency}
We next evaluate the end-to-end efficiency of PRISM when applied to Best-of-N sampling. Unlike pixel-based reward models, which require all N candidates to complete the full denoising trajectory and undergo VAE decoding before evaluation, PRISM can directly rank candidates from intermediate noisy latents. Fig. 6 plots the VBench Total against the inference time normalized to the VideoScore2-based BoN baseline. Selecting at Step 1 reduces the normalized time cost to 10\% on CogVideoX-2B and 19\% on Wan2.1-1.3B, a speedup of up to 10×, although selection at such a high noise level is close to the Random baseline on CogVideoX-2B. Slightly later selection already surpasses the strongest pixel-space baseline on each generator. With the larger PRISM (Wan2.1-14B), selection at Step 1 already outperforms all baselines. Selecting later yields only marginal gains on Wan2.1-1.3B and none on CogVideoX-2B, indicating that reliable candidate selection can be achieved well before the end of the denoising trajectory.

To further understand where these efficiency gains originate, we decompose the total inference latency into three major components: \textit{Denoise Generation}, \textit{VAE Decode}, and \textit{Reward Model} inference. As illustrated in \cref{fig:time_breakdown}, conventional pixel-based VRMs incur substantial costs in all three stages. They require complete denoising trajectories for all $N$ candidates, followed by VAE decoding of every candidate before reward evaluation. In addition, large VLM-based reward models introduce non-negligible evaluation latency, which is particularly pronounced for VideoScore2.

PRISM reduces the cost of each component through early latent-space selection. Candidates with low predicted rewards can be terminated before completing the remaining denoising steps, directly reducing the \textit{Denoise Generation} cost. Since the discarded candidates never need to be reconstructed in pixel space, VAE decoding is performed only for the final selected output rather than for all $N$ candidates. Finally, the lightweight query-based reward head introduces only marginal evaluation overhead compared with VLM-based VRMs. Together, these reductions explain the substantial end-to-end speedup observed in \cref{fig:time_breakdown} and demonstrate the practical efficiency of PRISM for video inference-time scaling.

\begin{table}[!t]
\centering
\caption{Comparison of model parameter scales between existing VLM-based reward models and PRISM variants. PRISM introduces negligible trainable parameters by freezing the DiT backbone.}
\label{tab:param_scale}
   \begin{tabular}{l r}
      \hline
      \textbf{Method} & \textbf{Param (M)} \\
      \hline
      VideoReward \cite{videoReward} & 2282.42 \\
      UnifiedReward \cite{unifiedreward} & 8027.35 \\
      VideoScore2 \cite{videoscore2} & 8292.16 \\
      \hline
      PRISM (CogVideoX-2B) & 709.67 \\
      \quad \textit{-- Trainable} & \textit{21.75} \\
      \hdashline
      PRISM (Wan2.1-1.3B) & 604.76 \\
      \quad \textit{-- Trainable} & \textit{21.80} \\
      \hdashline
      PRISM (Wan2.1-14B) & 4470.91 \\
      \quad \textit{-- Trainable} & \textit{21.80} \\
      \hline
   \end{tabular}
\end{table}

\begin{figure*}[!t]
\centering
\includegraphics[width=\textwidth]{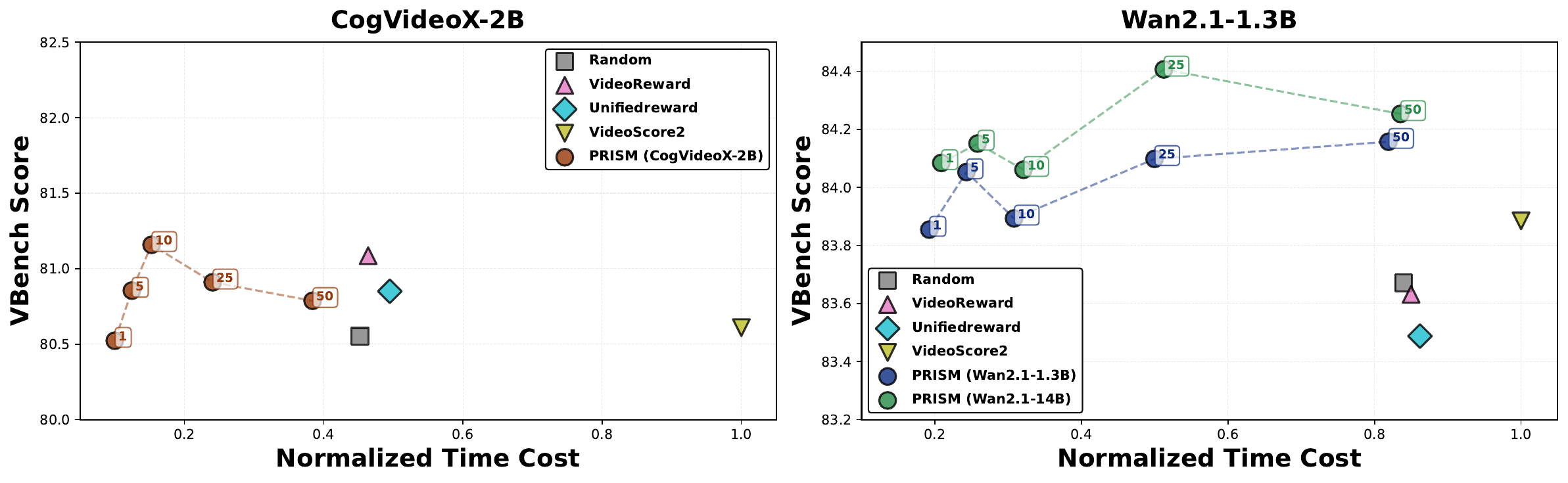}
\caption{Efficiency--quality trade-off of Best-of-N sampling on CogVideoX-2B (left) and Wan2.1-1.3B (right). The x-axis is the inference time normalized to the VideoScore2 baseline (1.0), and the y-axis is the VBench Total; numbers on PRISM markers denote the denoising step at which selection is performed. By selecting candidates at early denoising steps, PRISM surpasses the strongest pixel-space baseline on each generator at a fraction of its inference cost.}
\label{fig}
\end{figure*}

\begin{figure*}[!t]
\centering
\includegraphics[width=\textwidth]{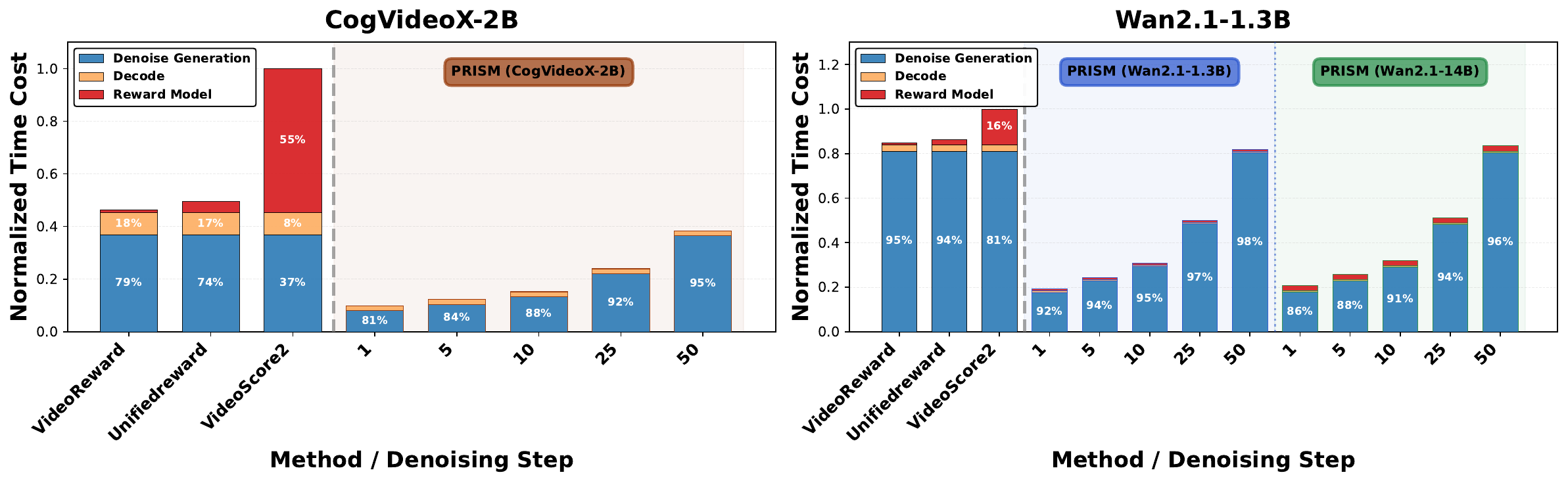}
\caption{Detailed breakdown of inference time cost during Best-of-N sampling. The total time is decomposed into Denoise Generation (blue), VAE Decode (orange), and Reward Model inference (red). Baseline methods (left of the dashed line) incur heavy costs across all three stages, requiring full denoising, full decoding for all candidates, and expensive VLM evaluations. In contrast, PRISM (right of the dashed line) substantially reduces total latency by truncating denoising trajectories, decoding only the selected candidate, and introducing minimal reward evaluation overhead. For PRISM, the x-axis denotes the intervention step during inference, where smaller values correspond to earlier high-noise states.}
\label{fig:time_breakdown}
\end{figure*}

\subsection{Ablation}

\subsubsection{Analysis of Full-Backbone FT vs. Frozen Backbone}

To validate the architectural necessity of PRISM's frozen backbone, we compare it against a Full-Backbone FT baseline using Wan2.1-1.3B. As shown in \cref{table:tuning_comparison}, while full fine-tuning marginally improves in-domain accuracy on VLRM-Bench (78.41 vs. 77.86), it suffers from out-of-domain overfitting on VideoGen-RewardBench, dropping to 63.22. Freezing the backbone acts as an essential regularizer that preserves pre-trained structural layouts and multi-modal priors. 

Furthermore, this design choice unlocks significant system-level efficiency. Unlike standalone evaluation frameworks that require massive model weight replication and redundant forward passes, PRISM grafts preference decoding directly into the active generation pipeline. Reusing intermediate features eliminates extra extraction and decoding overhead, providing a scalable and efficient path for model self-evolution.

\subsubsection{Ablation on RM Training Strategy for RL Alignment}

To evaluate the impact of our RM training strategy on policy optimization, we perform an ablation study comparing PRISM against a Full-Backbone FT baseline and the unaligned Base Model. We adopt a widely used few-step RL framework \cite{awm} on Wan2.1-1.3B \cite{wan} for stable and efficient training. We employ VBench for evaluation.

To provide implementation details for our training-time RL alignment, we deploy the training pipeline within the Flow-Factory framework \cite{flowfactory} on the Wan2.1-1.3B video model. We randomly sample 20,000 text prompts from VidProM \cite{vidprom} for training and reserve a disjoint set of 500 prompts for evaluation. All videos are generated at a resolution of $480 \times 832$ with 41 frames using the 4-NFE few-step Wan2.1-1.3B generator \cite{wan}, and are evaluated using VBench \cite{vbench} for fine-grained, multi-dimensional assessment. To ensure training stability and mitigate reward hacking during policy optimization, our off-policy formulation leverages a hybrid data mixture, combining online samples generated via an Exponential Moving Average (EMA) of the behavior policy with offline regularization data.

\begin{figure*}[!t]
  \centering
  \includegraphics[width=\textwidth]{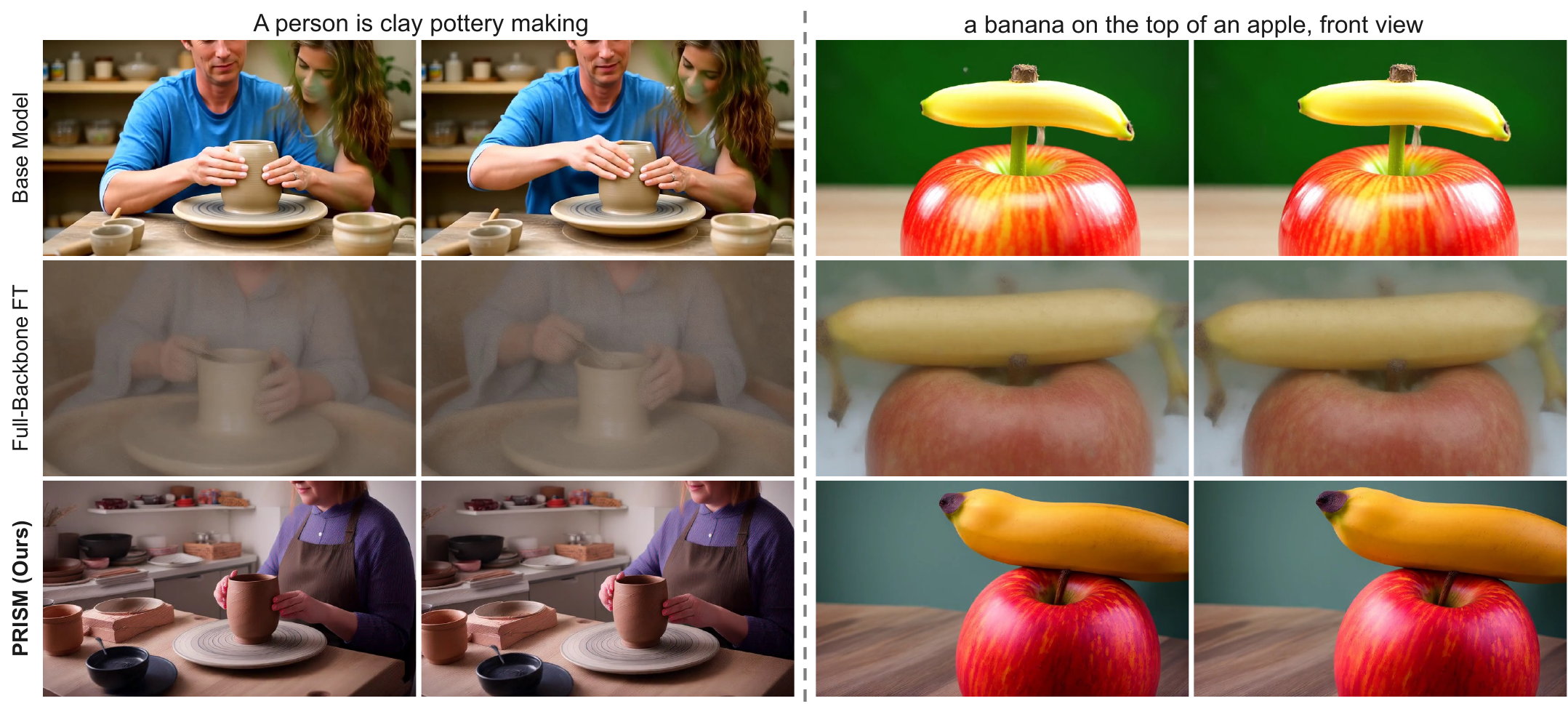}
    \caption{Qualitative comparison of training-time RL alignment on Wan2.1-1.3B.
    For each prompt, two frames of the generated video are shown.
    Base Model: the unaligned generator exhibits structural artifacts, e.g.,
    a spurious, semi-transparent second person (left) and a banana pierced by the
    apple stem (right).
    Full-Backbone FT: RL guided by the fully fine-tuned reward model is prone
    to reward hacking and degenerates into blurry, washed-out frames.
    PRISM (Ours): RL guided by PRISM preserves the generative prior and
    yields sharp, structurally coherent videos.}
  \label{fig:rl_comp}
\end{figure*}

\begin{figure}[!t]
    \centering
    \includegraphics[width=\columnwidth]{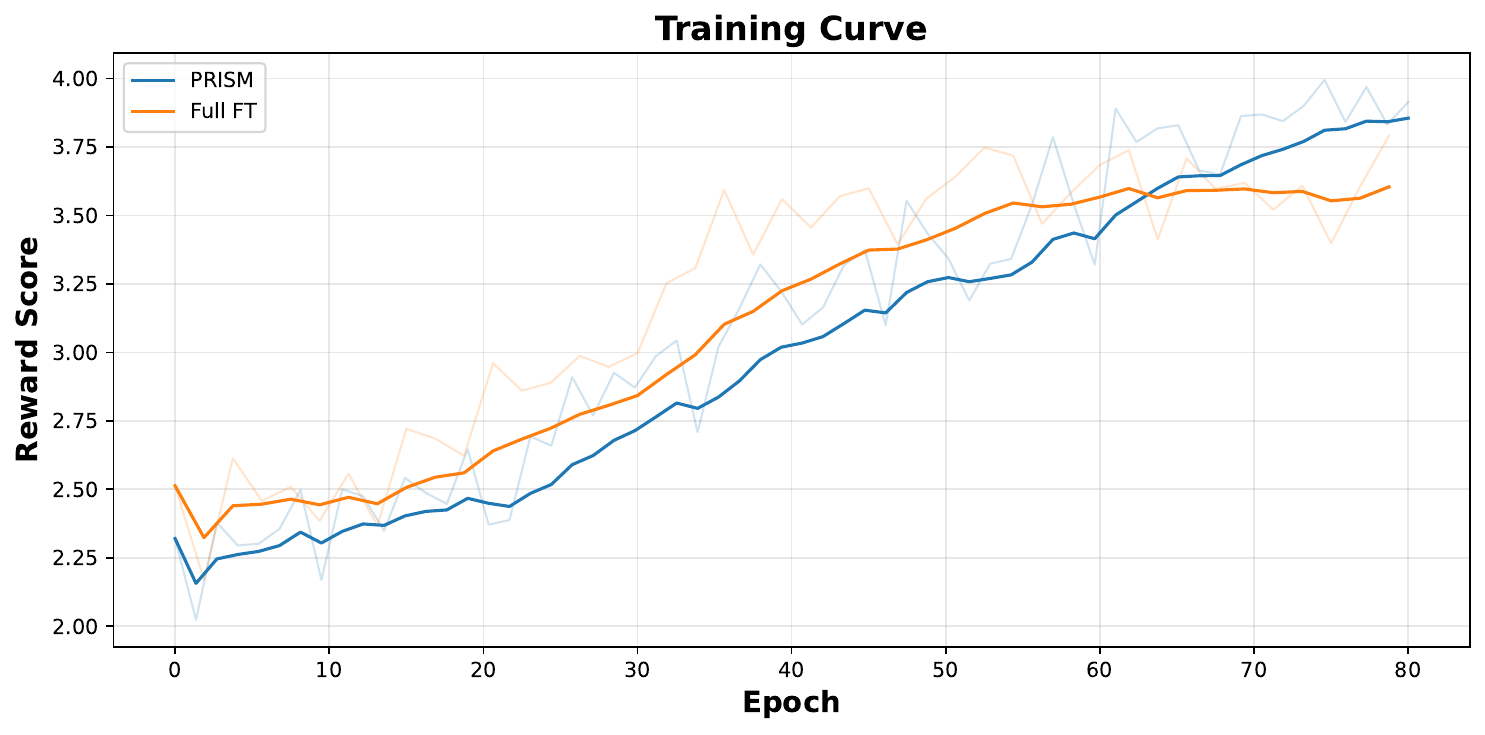} 
    \caption{Training curves of PRISM and Full-Backbone FT during RL alignment. Both strategies exhibit stable optimization trends and proper convergence on Wan2.1-1.3B.}
    \label{fig:training_curve}
\end{figure}

\begin{table}[!t]
\centering
\caption{Comparison between full fine-tuning and PRISM (Wan2.1-1.3B) on in-domain (VLRM-Bench) and out-of-domain (VideoGen-RewardBench) benchmarks. PRISM achieves stronger out-of-domain generalization by preserving pre-trained generative priors.}
\label{table:tuning_comparison}
   \begin{tabular}{ l c c }
      \noalign{\hrule}
      % \textbf{Method} & \makecell{\textbf{In-Domain}\\(VLRM-Bench)} & \makecell{\textbf{Out-of-Domain}\\(VideoGen-RewardBench)} \\
      \textbf{Method} & \textbf{In-Domain} & \textbf{Out-of-Domain} \\
      \noalign{\hrule}
      Full-Backbone FT & 78.41 & 63.22 \\
      PRISM (Wan2.1-1.3B)     & 77.86 & 72.93 \\
      \noalign{\hrule}
   \end{tabular}
\end{table}

\begin{table}[!t]
\centering
\caption{Quantitative results for Training-time Alignment (RL). We evaluate downstream RL alignment performance on Wan2.1-1.3B guided by different RM variants.}
\label{table_rl_1}
   \begin{tabular}{ l c c c }
      \hline
      \multirow{2}*{\textbf{Method}} & \multicolumn{3}{c}{\textbf{VBench}} \\
      ~ & Quality & Semantic & Total \\
      \hline
       Base Model & 83.2858 & 61.2937 & 78.8874 \\
       Full-Backbone FT & 74.5917 & 43.3329 & 68.3400 \\
       PRISM (Wan2.1-1.3B) & 85.5363 & 62.4471 & \textbf{80.9184} \\
      \hline
   \end{tabular}
\end{table}

As shown in \cref{table_rl_1}, RL alignment guided by the Full-Backbone FT strategy shows notable performance trade-offs, leading to lower VBench scores compared to the Base Model. Importantly, the training curves shown in \cref{fig:training_curve} indicate stable convergence despite the degraded downstream performance, suggesting that the performance drop is not caused by optimization failure. This indicates that an unconstrained, fully fine-tuned RM backbone lacks sufficient robustness, making the RL optimization susceptible to reward hacking—where the policy over-fits to non-essential preference features at the expense of generative priors. As illustrated in \cref{fig:rl_comp}, the generator aligned with the Full-Backbone FT reward model degenerates into blurry, washed-out frames that lose most scene detail. In contrast, our PRISM design provides a more robust reward signal, effectively preserving pre-trained structural priors while delivering superior downstream performance. Specifically, it improves over the Base Model on all three VBench scores, and in the examples of \cref{fig:rl_comp} it avoids the structural artifacts of the Base Model.

\begin{table*}[!t]
\centering
\caption{Ablation of feature aggregation: global pooling vs. query-based aggregation. Accuracy (\%) on VideoGen-RewardBench; the best results are in bold.}
\label{abla_table_1}
   \begin{tabular}{ l c c c c c c c c }
      \hline
      \multirow{2}*{Method} & \multicolumn{8}{c}{Timestep ($t$)} \\
      ~ & 989 & 949 & 899 & 699 & 499 & 299 & 99 & 0(Clean) \\
      \hline
      \textbf{CogVideoX-2B} \\
      \multicolumn{9}{l}{w/ Ties} \\
      Pool Agg + MLP & 38.77 & 45.68 & 47.50 & 50.51 & 51.43 & 51.47 & 51.51 & \textbf{52.14} \\
      Q-based Agg + MLP & \textbf{43.25} & \textbf{48.54} & \textbf{50.25} & \textbf{52.15} & \textbf{52.36} & \textbf{52.53} & \textbf{52.61} & 51.44 \\
      \hdashline
      \multicolumn{9}{l}{w/o Ties} \\
      Pool Agg + MLP & 43.79 & 53.27 & 55.74 & 59.26 & 60.51 & 60.67 & 60.89 & \textbf{62.15} \\
      Q-based Agg + MLP & \textbf{51.50} & \textbf{58.24} & \textbf{60.31} & \textbf{62.64} & \textbf{62.89} & \textbf{63.08} & \textbf{63.19} & 61.80 \\
      \hline
      \hline
      \textbf{Wan2.1-1.3B} \\
      \multicolumn{9}{l}{w/ Ties} \\
      Pool Agg + MLP & 43.67 & 52.62 & 54.64 & 55.68 & 56.09 & 56.13 & 56.24 & 55.60 \\
      Q-based Agg + MLP & \textbf{49.60} & \textbf{56.26} & \textbf{58.28} & \textbf{60.50} & \textbf{61.13} & \textbf{61.99} & \textbf{62.07} & \textbf{60.76} \\
      \hdashline
      \multicolumn{9}{l}{w/o Ties} \\
      Pool Agg + MLP & 50.38 & 61.76 & 64.25 & 64.83 & 65.21 & 65.22 & 65.35 & 65.15\\
      Q-based Agg + MLP & \textbf{59.56} & \textbf{67.56} & \textbf{70.01} & \textbf{72.68} & \textbf{73.43} & \textbf{74.42} & \textbf{74.51} & \textbf{72.93} \\
      \hline
   \end{tabular}
\end{table*}

\subsubsection{Impact of Feature Alignment and Aggregation}
To verify the effectiveness of our query-based aggregation mechanism, we conduct a comparative analysis against the baseline design of global adaptive pooling. Global pooling is a common yet rigid approach that collapses spatial and temporal dimensions via simple averaging, which often leads to the dilution of fine-grained preference signals—such as localized motion artifacts or subtle texture inconsistencies.

\Cref{abla_table_1} shows the ablation results on both CogVideoX-2B and Wan2.1-1.3B backbones. These results demonstrate that our query-based aggregation consistently outperforms the pooling baseline at all noisy timesteps ($t>0$) on both backbones; the only exception is the clean latent ($t=0$) on CogVideoX-2B, where pooling is marginally better. Specifically, the learnable queries $Q$ interact with the spatio-temporal features via cross-attention, allowing the model to dynamically focus on discriminative regions rather than treating all tokens with equal importance. This advantage is particularly evident on the Wan2.1-1.3B backbone, where the query-based aggregation maintains clearly higher accuracy at every timestep, even as $t$ increases. These findings validate our hypothesis that query-based information extractors can effectively preserve core preference information while mitigating the information loss inherent in straightforward dimensionality reduction.

\subsubsection{Ablation Analysis of Key Parameters}

We provide an ablation study on the key training hyperparameters of PRISM, namely the backbone feature layer index $N_b$, the query token count $N_q$, and the BTT loss threshold $\eta$. In each setting, one hyperparameter is varied while the others are kept at their default values ($N_b=12$, $N_q=1$, $\eta=5$). As shown in \cref{table:abla}, increasing $N_b$ to 15 provides only marginal accuracy gains at the expense of higher training costs, while decreasing it to 5 severely degrades performance. This indicates that sufficiently deep backbone features are essential for reliable preference prediction, whereas the benefit saturates beyond the default depth. For $N_q$ and $\eta$, the accuracy varies little over all tested values, showing that PRISM is robust to these choices, and our default configuration achieves the best or comparable accuracy across all evaluated timesteps.

\subsection{In-depth Analysis and Interpretability}

\begin{figure*}[!t]
  \centering
  \includegraphics[width=\textwidth]{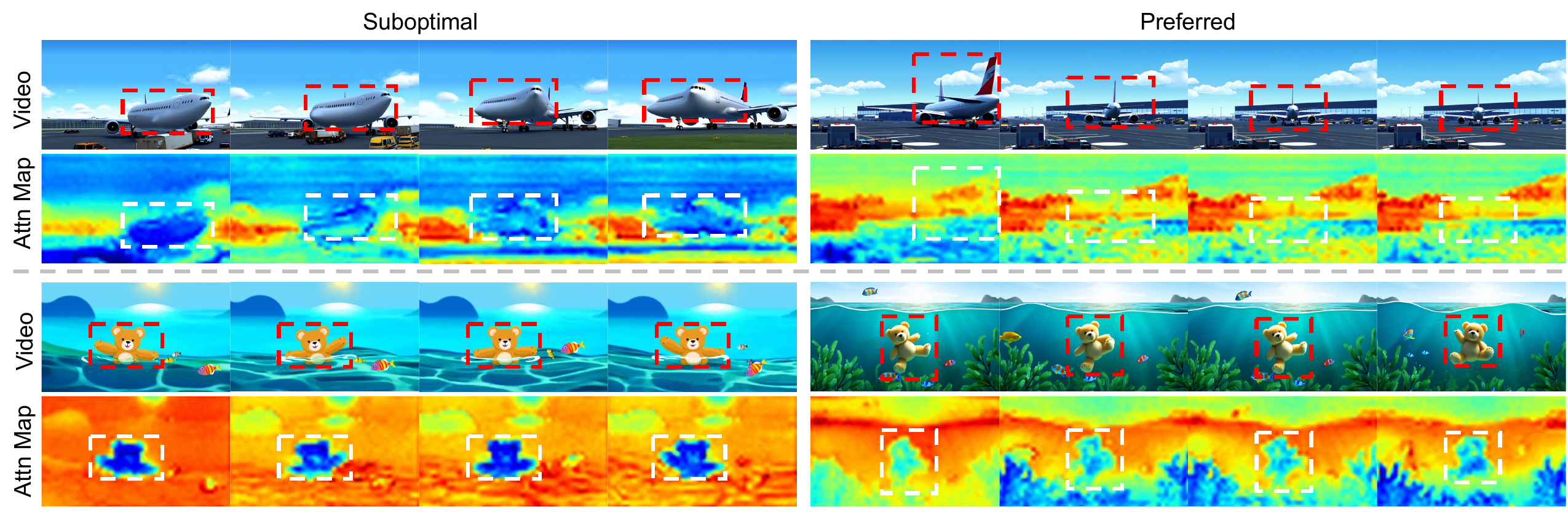}
    \caption{Comparative visualization of attention maps in the Query-based Aggregation module.
    We compare the cross-attention scores for suboptimal (left) and preferred (right) video samples. 
    As highlighted by the red boxes in the videos and white boxes in the attention maps, within the primary semantic regions (e.g., the aircraft or the teddy bear), the model assigns higher attention weights to high-fidelity structures. Conversely, regions containing structural distortions or malformed textures exhibit suppressed responses. This demonstrated sensitivity to generative quality, especially within core semantic areas, provides a highly interpretable basis for the model's preference assessment.}
  \label{fig:attn_vis}
\end{figure*}

We further investigate the rationale behind PRISM’s performance by visualizing the cross-attention mechanisms within the Query-based Aggregation head. As illustrated in \cref{fig:attn_vis}, the attention maps reveal a clear correlation between attention intensity and the structural integrity of \textit{semantic concept regions}.

Specifically, we observe that the learnable queries function as a quality-conditional filter. Within the targeted object regions (marked by red/white boxes), the attention intensity varies with generative fidelity. In the suboptimal samples where videos suffer from warped geometry or perceptual artifacts—such as the distorted aircraft fuselage or the malformed teddy bear—the attention scores are relatively low (appearing as cooler, blue regions). In contrast, the corresponding semantic regions in preferred samples elicit stronger and more focused responses. 

This comparative behavior demonstrates that PRISM does not merely aggregate global features; instead, its attention mechanism is discriminatively sensitive to the ``health'' of the generated content. By assigning higher weights to high-fidelity semantic signals while discounting regions with localized structural distortions, the module addresses the challenge of information dilution, providing an interpretable foundation for latent-space preference alignment.

\begin{table}[!t]
\centering
\caption{Ablation of key hyperparameters on Wan2.1-1.3B. Each row changes one of $N_b$, $N_q$, and $\eta$ from the default setting (``Ours''). Accuracy (\%, w/o Ties) on VideoGen-RewardBench.}
\label{table:abla}
   \begin{tabular}{ l c c c c }
      \hline
      \multirow{2}*{\textbf{Method}} & \multicolumn{4}{c}{\textbf{Timestep ($t$)}} \\
      ~ & 899 & 499 & 299 & 0(Clean) \\
      \hline
      $N_b=5$ & 66.10 & 69.27 & 69.31 & 66.85 \\
      $N_b=15$ & 70.59 & 74.06 & 75.09 & 73.42 \\
      \hline
      $N_q=4$ & 70.28 & 73.41 & 74.29 & 72.82 \\
      $N_q=8$ & 70.15 & 73.35 & 73.79 & 72.14 \\
      \hline
      $\eta=3$ & 69.96 & 73.31 & 74.18 & 72.58 \\
      $\eta=8$ & 69.70 & 73.28 & 73.81 & 72.55 \\
      \hline
      Ours & 70.01 & 73.43 & 74.42 & 72.93 \\
      \hline
   \end{tabular}
\end{table}

\section{Limitations and Future Work}

The primary constraint of PRISM lies in its architectural coupling with the underlying diffusion backbone. Specifically, the effectiveness of our reward head relies on the specific latent representations learned by a particular VAE. This necessitates that both the evaluator and the generator reside in the same latent domain. When applying PRISM to evaluate outputs from a generative model with a different VAE design, the latent features must be decoded into pixel space and subsequently re-encoded into the evaluator’s latent space. This additional computational overhead limits the ``plug-and-play'' capability of PRISM across heterogeneous model families. This coupling is at the level of the latent space rather than the specific model: generators that share a VAE can directly use a PRISM trained on a different backbone. For example, PRISM (Wan2.1-14B) guides the Wan2.1-1.3B generator in \cref{main_table_2} and achieves the best BoN result.

Despite this limitation, PRISM currently serves as a highly efficient, specialized ``expert evaluator'' for specific model lineages. Future work will explore the incorporation of latent-space alignment methods or cross-model adapters to achieve broader robustness and backbone-agnostic evaluation. Since PRISM provides a reward at any timestep without decoding, it could also serve as a dense, step-level signal for preference optimization.

\section{Conclusion}
We presented PRISM, an efficient latent-space video reward framework that bridges human preferences and video generation. By repurposing noise-resilient spatiotemporal priors from frozen Video Diffusion Transformers, PRISM circumvents the massive computational overhead and noise-sensitivity of traditional pixel-based models. Our study demonstrates these generative priors offer a robust foundation for preference learning across diverse architectures (e.g., CogVideoX and Wan2.1). Technically, our Query-based Aggregation extracts preference signals from high-dimensional latents, while attention analysis reveals its interpretability in suppressing regional artifacts. Extensive evaluations confirm PRISM achieves the best preference accuracy on noisy latents. By enabling early-stage selection, PRISM significantly reduces latency while matching or exceeding pixel-space baselines in generation quality. Ultimately, PRISM provides a practical alignment tool and advances our understanding of the evaluative capabilities of generative models.

{
\bibliographystyle{IEEEtran}
\bibliography{reference}
}

\begin{IEEEbiography}[{\includegraphics[width=1in,height=1.25in,clip,keepaspectratio]{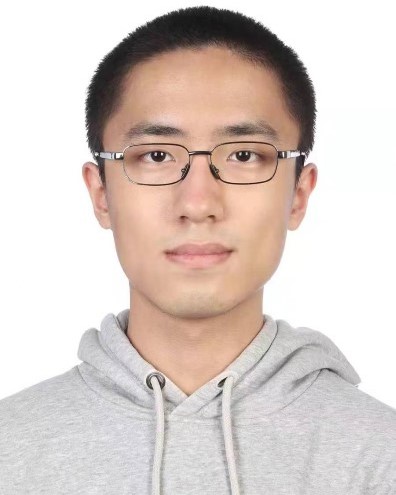}}]{Haoxuan Wu} received the B.Eng. degree in computer science and technology from North China Electric Power University, Beijing, China, in 2022, and the M.S. degree in multimedia information technology from the City University of Hong Kong, Hong Kong, in 2023, where he is currently pursuing the Ph.D. degree with the Department of Electrical Engineering. His research interests include deep learning and computer vision.
\end{IEEEbiography}

\vspace{-10pt}

\begin{IEEEbiography}[{\includegraphics[width=1in,height=1.25in,clip,keepaspectratio]{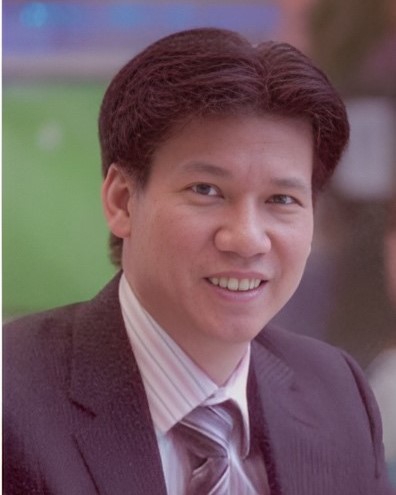}}]{Lai-Man Po} (Senior Member, IEEE) received the B.S. and Ph.D. degrees in electronic engineering from the City University of Hong Kong, Hong Kong, in 1988 and 1991, respectively. 

He has been with the Department of Electronic Engineering, City University of Hong Kong, since 1991, where he is currently an Associate Professor of the Department of Electrical Engineering. He has authored over 150 technical journals and conference papers. His research interests include image and video coding with an emphasis on deep learning-based computer vision algorithms. 
He is a member of the Technical Committee on Multimedia Systems and Applications and the IEEE Circuits and Systems Society. He was the Chairman of the IEEE Signal Processing Hong Kong Chapter in 2012 and 2013. He was an Associate Editor of HKIE Transactions from 2011 to 2013. He also served on the Organizing Committee, of the IEEE International Conference on Acoustics, Speech and Signal Processing in 2003, and the IEEE International Conference on Image Processing in 2010.
\end{IEEEbiography}

\begin{IEEEbiography}[{\includegraphics[width=1in,height=1.25in,clip,keepaspectratio]{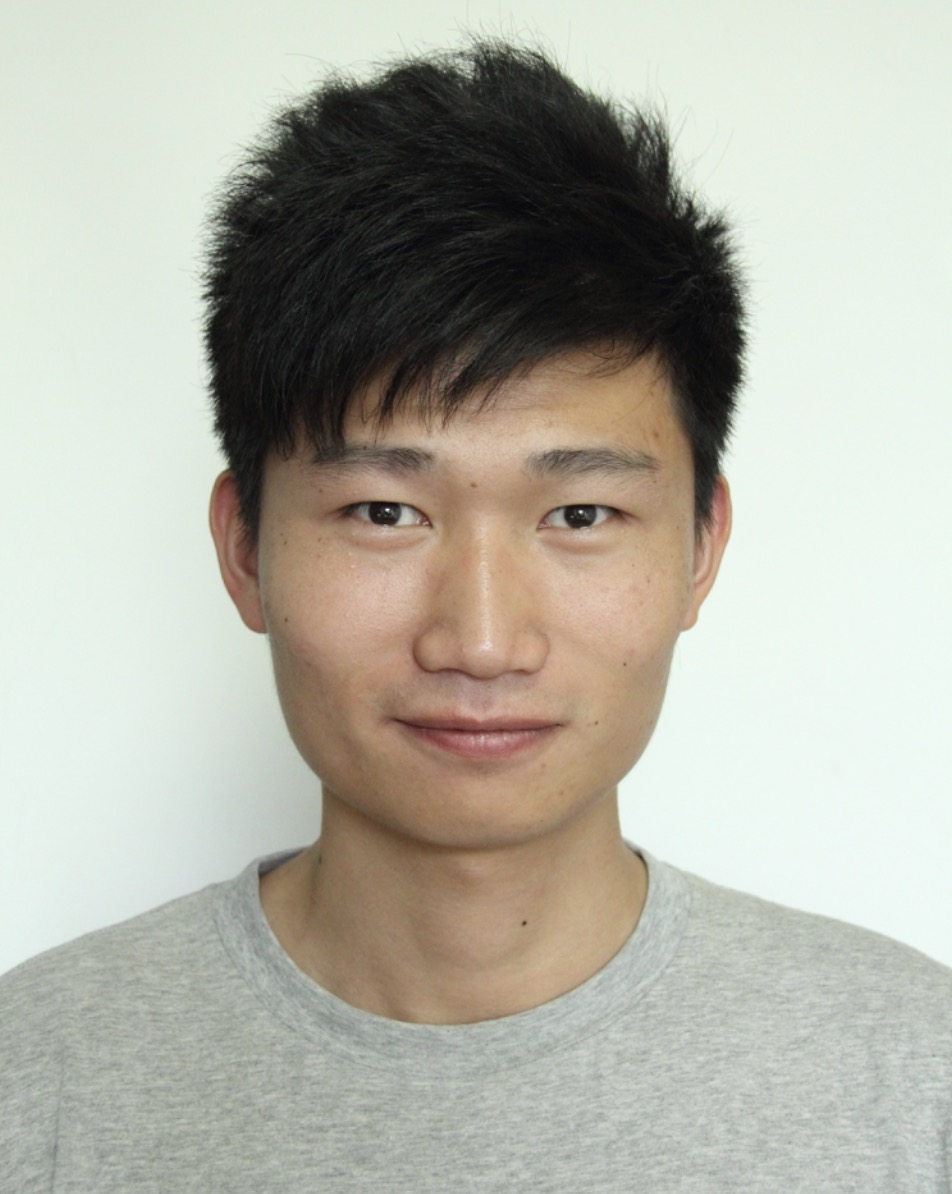}}]{Mengyang Liu} is currently a Research Scientist with Video Rebirth, focusing on foundation video model research. He was previously a Research Engineer with Tencent Video, Tencent Holdings Ltd. His research interests include image and video processing, video/image generation, video embedding and retrieval. He received the B.Eng. degree in optoelectronic engineering from the Shanghai University of Electric Power, Shanghai, China, in 2014, and the M.Sc. degree in electronic and information engineering and the Ph.D. degree from the City University of Hong Kong, in 2015 and 2019, respectively.
\end{IEEEbiography}

\begin{IEEEbiography}[{\includegraphics[width=1in,height=1.25in,clip]{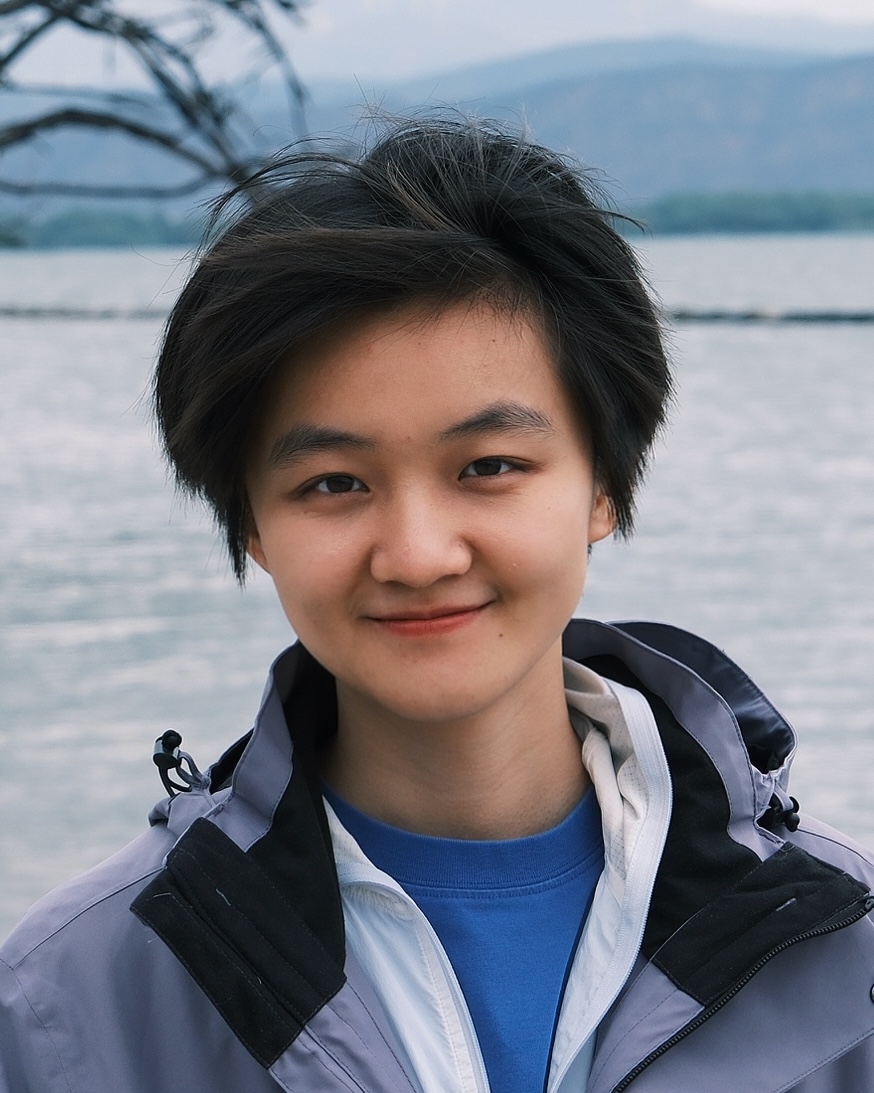}}]{Kun Li} (Graduate Student Member, IEEE) received the B.Eng. degree in IoT Engineering from the University of Electronic Science and Technology of China, Chengdu, China, in 2022. She is currently pursuing the Ph.D. degree with the Department of Electrical Engineering from the City University of Hong Kong. Her research interests include computer vision, generative models, and multimodal large language models.
\end{IEEEbiography}

\begin{IEEEbiography}[{\includegraphics[width=1in,height=1.25in,clip,keepaspectratio]{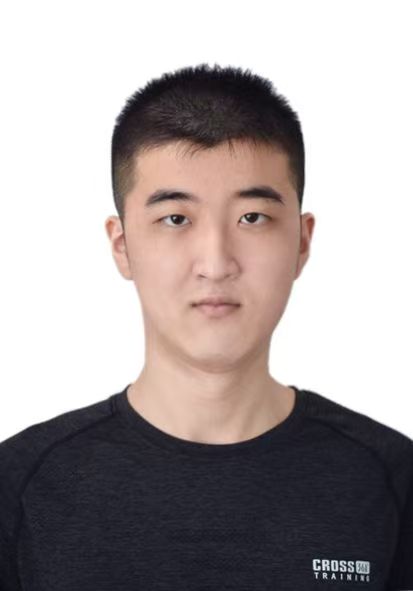}}]{Hongzheng Yang}
received the B.Eng. degree in Artificial Intelligence from Beihang University in 2022.
He is currently pursuing the Ph.D. degree at The Chinese University of Hong Kong. His research interests include AI alignment, Video generation and AI for social good (such as healthcare applications).
He has authored 5 first-authored works and has contributed to over 10 papers in prestigious venues, including ICML, TIP, NeurIPS, TMI, CVPR, and Nature Communications.
\end{IEEEbiography}

\begin{IEEEbiography}[{\includegraphics[width=1in,height=1.25in,clip,keepaspectratio]{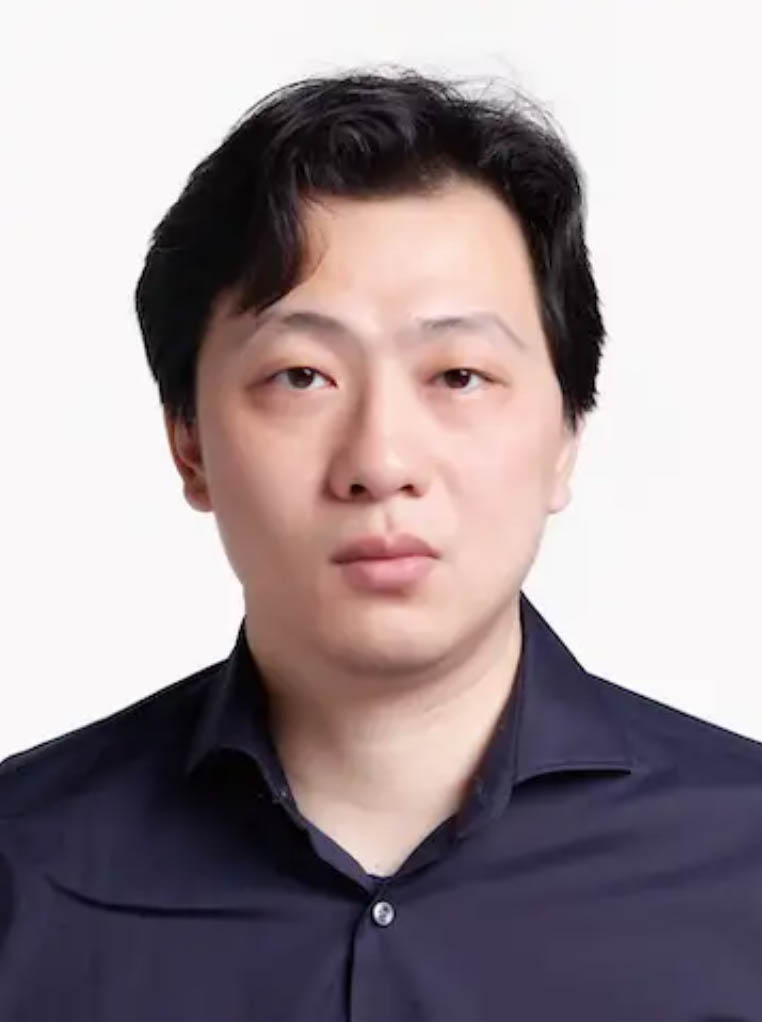}}]{Dr. Wei Liu}
is currently the Founder and CEO at Video Rebirth, a world model company. He had been a Distinguished Scientist of Tencent in 2016--2024, taking the director of Multimodal AI at Tencent Data Platform and the director of Computer Vision at Tencent AI Lab. Dr. Liu was the core co-founder and chief AI scientist of Tencent's generative foundation model series--HunYuan, now named Tencent HY. Dr. Liu is an Adjunct Professor at City University of Hong Kong, a Guest Full Professor at Fudan University, China, and a Guest Full Professor at Zhejiang University, China. Prior to joining Tencent, he received the Ph.D. degree in EECS from Columbia University, USA, and was a research scientist and staff member of IBM T. J. Watson Research Center, USA.

Dr. Liu has long been devoted to fundamental research and technological development in core fields of AI. His research works win numerous awards and honors, e.g., the 2011 Facebook Fellowship, the 2012 IBM Josef Raviv Memorial Postdoctoral Fellowship, the 2013 Jury Award for Best Thesis of Columbia University, the 2016 and 2017 SIGIR Best Paper Award Honorable Mentions, the 2018 ``AI's 10 To Watch'' Award, the 2020--2025 World's Top 2\% Scientists honor, etc. Dr. Liu is an Associate/Action Editor of internationally leading AI\&CS journals IEEE Transactions on Pattern Analysis and Machine Intelligence (TPAMI), ACM Computing Surveys, and Transactions on Machine Learning Research (TMLR), and is an Area Chair of top-tier computer science and AI conferences NeurIPS, ICML, ICLR, CVPR, ICCV, ECCV, IJCAI, and AAAI. Dr. Liu is a Fellow of the AAAS (American Association for the Advancement of Science), IEEE (Institute of Electrical and Electronics Engineers), and IAPR (International Association for Pattern Recognition), and is an Elected Member of the NAAI (National Academy of Artificial Intelligence) and ISI (International Statistical Institute). Dr. Liu is an Elected Member of the IEEE Fellow Committee for 2024--2025.
\end{IEEEbiography}

\end{document}